\documentclass{article}

    \PassOptionsToPackage{numbers, compress}{natbib}

    \usepackage[final]{neurips_2024}

\usepackage[utf8]{inputenc} 
\usepackage[T1]{fontenc}    
\usepackage{hyperref}       
\usepackage{url}            
\usepackage{booktabs}       
\usepackage{amsfonts}       
\usepackage{nicefrac}       
\usepackage{microtype}      
\usepackage{xcolor}         
\usepackage{colortbl}
\usepackage{amsmath,amssymb,amsfonts}
\usepackage{graphicx}
\usepackage{subfigure}
\usepackage{multirow}
\usepackage{listings}
\usepackage{pythonhighlight}
\usepackage{paralist}
\usepackage{arydshln}
\usepackage{pifont}
\usepackage{wrapfig}
\usepackage{caption}
\usepackage{natbib}

\definecolor{lightblue}{RGB}{50,150,220}

\title{Generalizing Weather Forecast to Fine-grained Temporal Scales via Physics-AI Hybrid Modeling}

\author{%
  Wanghan Xu\thanks{This work was done during his internship at Shanghai Artificial Intelligence Laboratory.} \\
  Shanghai Jiao Tong University \\
  Shanghai AI Laboratory \\
  \texttt{xu\_wanghan@sjtu.edu.cn} \\
  \And
  Fenghua Ling \\
  Shanghai AI Laboratory \\
  \texttt{lingfenghua@pjlab.org.cn} \\
  \And
  Wenlong Zhang \\
  Shanghai AI Laboratory \\
  \texttt{zhangwenlong@pjlab.org.cn} \\
  \And
  Tao Han \\
  Shanghai AI Laboratory \\
  \texttt{hantao.dispatch@pjlab.org.cn} \\
  \And
  Hao Chen \\
  Shanghai AI Laboratory \\
  \texttt{chenhao1@pjlab.org.cn} \\
  \And
  Wanli Ouyang \\
  Shanghai AI Laboratory \\
  \texttt{ouyangwanli@pjlab.org.cn} \\
  \And
  Lei Bai\thanks{Corresponding author.} \\
  Shanghai AI Laboratory \\
  \texttt{bailei@pjlab.org.cn}
}

\begin{document}

\maketitle

\begin{abstract}
Data-driven artificial intelligence (AI) models have made significant advancements in weather forecasting, particularly in medium-range and nowcasting. However, most data-driven weather forecasting models are black-box systems that focus on learning data mapping rather than fine-grained physical evolution in the time dimension. Consequently, the limitations in the temporal scale of datasets prevent these models from forecasting at finer time scales. This paper proposes a physics-AI hybrid model (i.e., WeatherGFT) which \textbf{G}eneralizes weather forecasts to \textbf{F}iner-grained \textbf{T}emporal scales beyond training dataset. Specifically, we employ a carefully designed PDE kernel to simulate physical evolution on a small time scale (e.g., 300 seconds) and use a parallel neural networks with a learnable router for bias correction. Furthermore, we introduce a lead time-aware training framework to promote the generalization of the model at different lead times. The weight analysis of physics-AI modules indicates that physics conducts major evolution while AI performs corrections adaptively. Extensive experiments show that WeatherGFT trained on an hourly dataset, effectively generalizes forecasts across multiple time scales, \textbf{\textit{including 30-minute, which is even smaller than the dataset's temporal resolution}}. Code is available at \textcolor{blue}{\textbf{https://github.com/black-yt/WeatherGFT}} .
\end{abstract}

\section{Introduction}

\textbf{Weather forecasting} plays a vital role in modern society, impacting a wide range of human activities. For example, minute-level precipitation nowcasting is particularly valuable for short-term planning, such as outdoor activities, while medium-range forecasts that offer daily predictions play a crucial role in long-term strategic decisions like maritime trade. This field has witnessed remarkable advancements in recent years, largely attributed to the rapid progress of machine learning-based (ML) weather forecasting models~\cite{ling2024artificial}, spanning from nowcasting to medium-range forecasts.

\textbf{Prior studies} tackle the problem of weather forecasting by leveraging data-driven models trained on benchmark weather forecasting datasets, such as WeatherBench~\cite{rasp2020weatherbench} and ERA5~\cite{hersbach2020era5}. Prevalent medium-range forecasting models (e.g., FourCastNet~\cite{kurth2023fourcastnet}, GraphCast~\cite{lam2023learning}, and FengWu~\cite{chen2023fengwu}) are commonly trained on the aforementioned hourly datasets to generate global forecasts with a time interval of 6-hour, can not offer finer predictions like 30-minute forecasts for nowcasting.

\textbf{A significant limitation} of current ML-based weather forecasting models~\cite{kurth2023fourcastnet,bi2023accurate,lam2023learning,chen2023fengwu,han2024weather} lies in their black-box training paradigm~\cite{verma2024climode, guidotti2018survey}, that is, primarily focusing on learning the mapping of data pairs with a fixed lead time (e.g., 6 hours), without explicitly incorporating the laws of atmospheric dynamics which govern finer-grained physical evolution processes. Consequently, this training paradigm brings a significant challenge for weather forecasting: \textbf{\textit{existing black-box AI models are unable to generalize at finer temporal scales beyond the inherent time resolution of the training datasets due to the absence of fine-grained physics modeling}}.

\textbf{To address this challenge}, we propose WeatherGFT, a physics-AI hybrid model capable of simulating weather changes on fine-grained time scales through a set of partial differential equations (PDEs)~\cite{strauss2007partial}. WeatherGFT consists of an encoder, multiple stacked \textbf{HybridBlocks} and a decoder. As the core of our model, HybridBlock contains two branches: One utilizes PDE kernels to conduct physical evolution over small time scales, while the other employs neural networks to learn unresolved atmospheric processes and perform bias correction on the physical evolution. These two branches are adaptively fused through a \textbf{learnable router} initialized as 0.5:0.5. Unlike existing models~\cite{kurth2023fourcastnet, lam2023learning, chen2023fengwu} trained with a fixed lead time, we introduce a \textbf{lead time-aware framework} through multi-lead time training strategy and a lead time conditional decoder~\cite{nguyen2023scaling, andrychowicz2023deep}, enabling the model to generalize to finer-grained temporal scales. Experiments demonstrate that our method is capable of forecasting at different lead times within one single model and one unified framework, overcoming the limitations of the dataset's temporal resolution and \textbf{\textit{enabling 30-minute forecasts with an hourly dataset}}.

\begin{wrapfigure}[18]{r}{0.52\textwidth}
    \vspace{-18pt}
    \includegraphics[scale=0.68, trim={0 0 0 0}, clip]{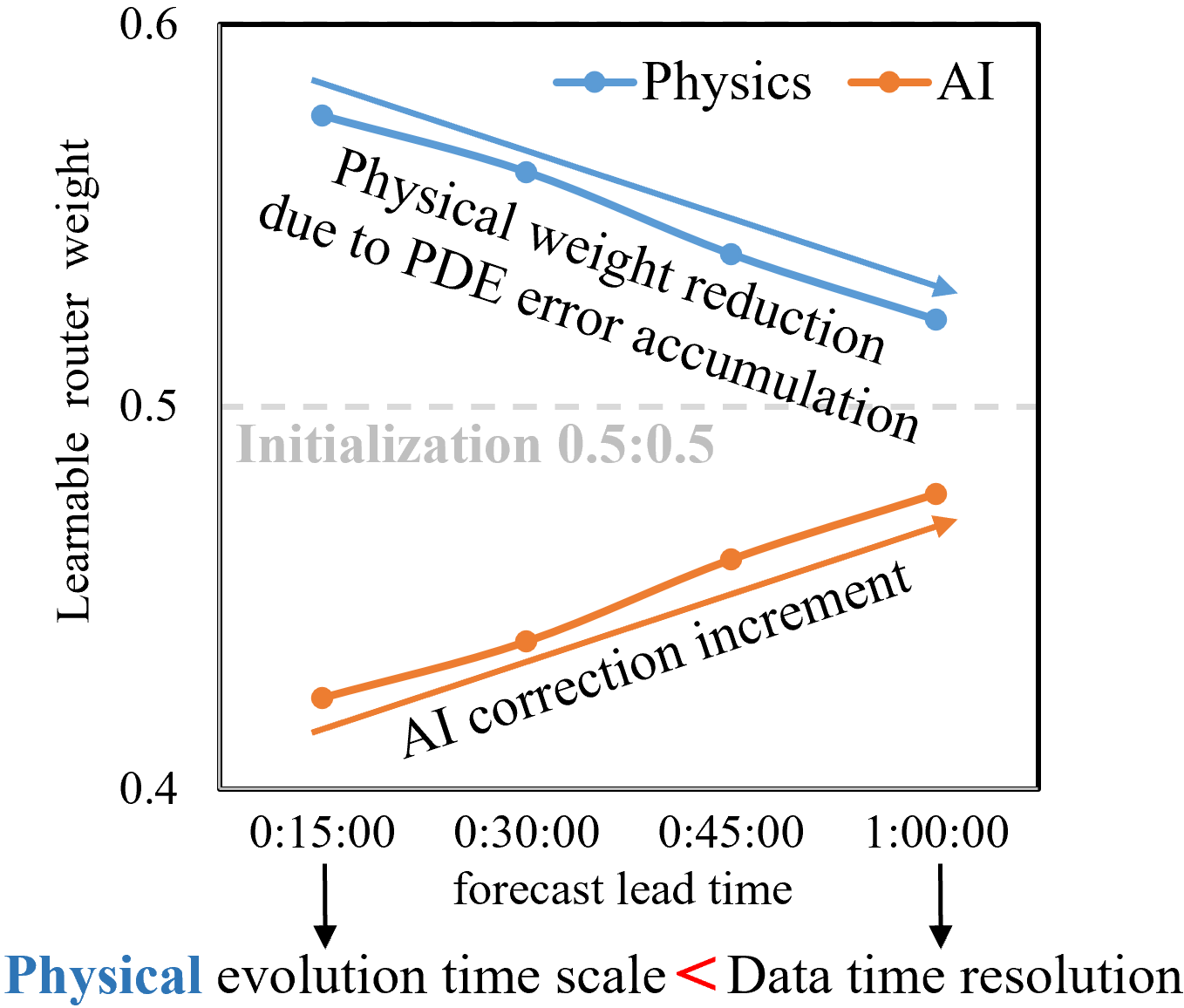}
    \vspace{-5pt}
    \caption{\textbf{Learnable router weight}. The role of \textcolor{lightblue}{\textbf{physics}} and \textcolor{orange!90}{\textbf{AI}} at different lead times: \textcolor{lightblue}{\textbf{major evolution}} and \textcolor{orange!90}{\textbf{adaptive correction}} (details in Sec. \ref{sec:weight}).}
    \label{fig:ratio_mean}
\end{wrapfigure}

Additionally, we find two interesting insights by examining the learnable route weight of the hybrid physical-AI modules at different lead times, as depicted in Figure \ref{fig:ratio_mean}: \textbf{a)} The physical weight is consistently higher than the AI, indicating the significant role played by the PDE kernel. \textbf{b)} As the lead time increases, the weight of AI gradually increases. We attribute this increment to the errors accumulation of PDE kernel during the evolution process, necessitating more AI corrections. In summary, when there is training data available at the lead time, such as at 1:00:00, the fitting ability of AI is enhanced. Conversely, at the lead time without training data, such as at 0:30:00, the importance of physical evolution becomes more pronounced, which confirms our motivation: \textbf{WeatherGFT can benefit from both physics and AI adaptively.}

We summarize the contributions of this paper as follows:
\begin{compactitem}
\label{sec:contributions}
\item We propose a physics-AI hybrid model that incorporates physical PDEs into the networks, enabling the simulation of fine-grained physical evolution through its forward process.
\item With the flexible PDE kernel and new lead time-aware training framework, our model performs multiple lead time forecasts, which bridges the nowcast and medium-range forecast.
\item For the first time, our model extends the forecasting ability learned from an hourly dataset to make accurate predictions at a finer time scale, i.e., 30 minutes.
\item Our model exhibits strong generalization ability while maintaining prediction errors comparable to those of pure AI and physical models.
\end{compactitem}

\section{Related Work}

\paragraph{Data-driven Weather Forecasting.} In recent years, data-driven weather forecasting models based on machine learning have developed rapidly~\cite{ben2024rise}, especially for medium-range weather forecast~\cite{wagner1989medium}, which provides weather variables for the next few days. Clare et al.\cite{clare2021combining} propose a weather forecasting approach using stacked ResNets~\cite{he2016deep}, but their model only considers geopotential and temperature, which is limited for real-world forecasting applications. FourCastNet~\cite{kurth2023fourcastnet} expands the model to include additional variables such as wind at different heights, and employs Adaptive Fourier Neural Operator (AFNO)~\cite{guibas2021adaptive} networks for prediction. Pangu-Weather~\cite{bi2023accurate} utilizes the 3D Swin Transformer~\cite{zhang2023spine} and introduces hierarchical temporal aggregation to minimize iterations in the autoregressive forecasting, followed by FengWu~\cite{chen2023fengwu, xiao2023fengwu}, FuXi~\cite{chen2023fuxi} and other Transformer-based~\cite{vaswani2017attention} prediction models. Apart from Transformers, GraphCast~\cite{lam2023learning} and Keisler~\cite{keisler2022forecasting} adopt a graph representation of the Earth and employ Graph Neural Network (GNN)~\cite{zhou2020graph} for weather prediction.

In addition to medium-range weather forecast, nowcast~\cite{browning1989nowcasting, wang2024predbench} is another important field in weather forecast, which usually provides 30-minute forecasting of severe convective weather like thunderstorms. OFAF~\cite{nie2021ofaf}, Preciplstm~\cite{ma2022preciplstm}, SimVP~\cite{gao2022simvp} use convolutions to capture spatial information and model temporal information through networks such as Long Short-Term Memory~\cite{yu2019review} or Recurrent Neural Network~\cite{yin2017comparative}. Earthformer~\cite{gao2022earthformer} and CasCast~\cite{gong2024cascast} use Transformer-based models for nowcasting. The former proposes cuboid attention to efficiently model space-time information, and the latter uses the diffusion model~\cite{song2020denoising} to address the problem of blur output. These nowcast models focus on minute-level forecasts for specific regions, and is difficult to forecast for long-term such as 5-day.


Consequently, there exists a significant gap (global vs. regional, day-level vs. minute-level, long-term vs. shot-term) between medium-range forecasts and nowcasts. Integrating AI models with physical guidance to make finer-grained predictions can bridge this gap.


\paragraph{Physical Neural Networks.} Most data-driven models commonly neglect the incorporation of physics and treat networks as black-boxes. In order to enhance the consistency of predictions with respect to physical laws, PINNs~\cite{cai2021physics}, PINO~\cite{li2021physics}, and DeepPhysiNet~\cite{li2024deepphysinet} add PDE loss to overall training loss. Nevertheless, these methods of changing loss functions often require balancing the weights between different PDEs, and the training results are heavily affected by hyperparameters. PI-HC-MoE~\cite{chalapathi2024scaling}, ClimODE~\cite{verma2024climode} integrate physical processes into the networks, but they do not explicitly simulate the physical evolution of distinct variables based on PDEs. Instead, they implement the evolution using general kernels, such as Euler kernels~\cite{vadyala2022physics}. NeuralGCM~\cite{kochkov2023neural} employs neural networks to parameterize a dynamic core. However, it is primarily designed for medium-range forecasting. These works typically focus on forecasting at fixed lead times, rather than leveraging physical laws to generalize to finer-grained time scales beyond the training datasets.

\section{Method}

\subsection{Problem Formulation}

Weather forecasting aims to predict future weather states $\mathcal X_{t}$ given current weather states $\mathcal X_{0}$:


\begin{equation}
    F_{\theta}(\mathcal X_{0})=P(\mathcal X_{t}|\mathcal X_{0})
\end{equation}


where $\theta$ represents the parameters of the model and $t$ denotes the lead time. The weather state $\mathcal X \in \mathbb{R}^{C \times H \times W}$ consists of $C$ atmospheric variables across different pressure levels. Each variable is characterized by an $H\times W$ matrix that corresponds to the projection of the Earth's plane. 

Assuming that the time resolution of the dataset is $t_{data}$, the lead time $t$ for data-driven models can only be equal to or greater than $t_{data}$, because these models are trained using data pairs $(\mathcal{X}_{0}, \ \mathcal{X}_{t})$ sampled from the dataset. Consequently, black-box AI models~\cite{kurth2023fourcastnet, bi2023accurate, lam2023learning, chen2023fengwu, han2024cra5, han2024fengwu, gong2024postcast} are unable to forecast at finer lead times such as $\frac{1}{2} t_{data}$, indicating \textbf{a lack of temporal generalization ability}.


\subsection{WeatherGFT Overview}

\begin{figure}[t]
\centerline
{\includegraphics[width=15cm]{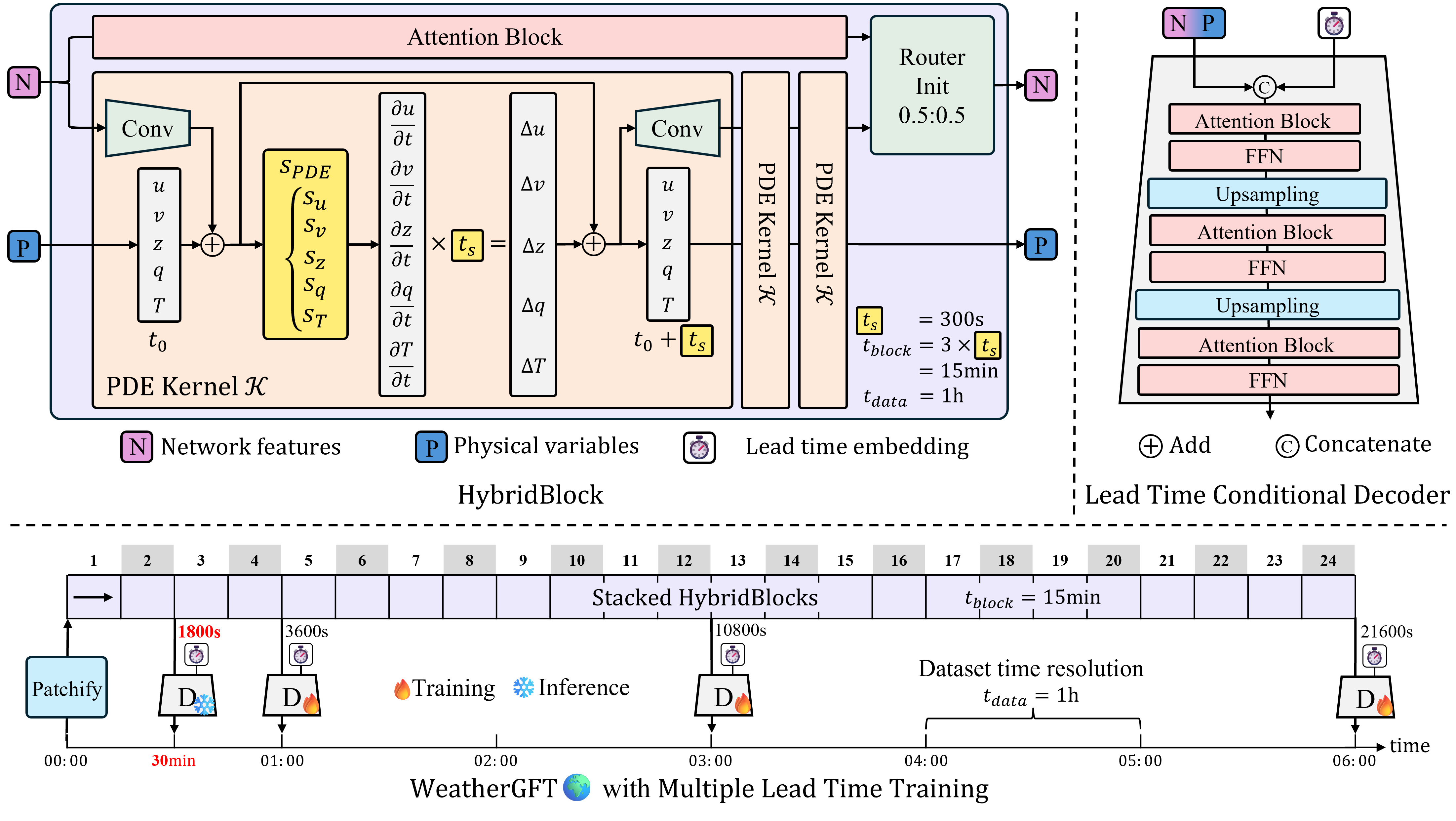}}
\caption{\textbf{Overview of WeatherGFT.} HybridBlock serves as the fundamental unit of the model, consisting of three PDE kernels, a parallel Attention Block, and a subsequent learnable router. A lead time conditional decoder is employed to generate forecasts for different lead times.}
\label{fig: Framework}
\end{figure}

As shown in Figure~\ref{fig: Framework}, our model consists of an encoder to patchify the weather states into tokens~\cite{vaswani2017attention}, multiple (specifically, 24) stacked HybridBlocks to preform weather evolution via PDE modeling, and a decoder to generate predictions under specific lead-time conditions.

Specifically, to enable our model to generalize at a finer-grained temporal resolution, we employ PDEs to model the evolution at a finer time scale:


\begin{equation}
\label{eq2}
    \mathcal{X}_{t_s} = \mathcal{K}(\mathcal{X}_{0}), \ \mathrm{where}\  t_s = \frac{1}{m} t_{data},\  m \in \mathbb{Z}^{+}
\end{equation}


We simulate the physical evolution from $\mathcal{X}_{0}$ to $\mathcal{X}_{t_s}$ through a uniquely designed PDE kernel (details in Section \ref{sec:pdekernel}), where $t_s$ is much smaller than the time resolution $t_{data}$ of the dataset, allowing model to capture fine-grained weather changes. By stacking PDE kernels $ \mathcal{K}$, the longer evolution can be achieved like $\mathcal X_{t_{data}} = \mathcal{K}_m  \dots \mathcal{K}_{2}\ \mathcal{K}_{1}\  \mathcal{X}_{0}$. In this paper, we set $m$ to 12, that is, $t_s=\frac{1}{12}t_{data}$.

To mitigate the issue of error accumulation as the number of evolutionary steps increases, we introduce a parallel Attention Block~\cite{vaswani2017attention} that performs bias correction for every 3 iterations of $\mathcal{K}$. Additionally, a learnable router initialized as $0.5:0.5$, is employed to adaptively fuse features from PDE kernels and the Attention Block. We encapsulate three PDE kernels $\mathcal{K}$ and one parallel Attention Block within a HybridBlock, whose evolution time is $t_{block} =3 \times t_s = \frac{1}{4}t_{data}$. 

Our model can not only forecast at lead times equal to or greater than $t_{data}$, but also generalize to finer-grained time scale such as $\frac{1}{2} t_{data}$ even in the absence of corresponding training data pairs. This is achieved by modeling the physical evolution of $t_{block} = \frac{1}{4}t_{data}$, rather than simply learning from data pairs $(\mathcal{X}_{0}, \mathcal{X}_{t_{data}})$ sampled from the dataset. Notably, these generalized finer-grained predictions of our model outperform temporal interpolation on multiple metrics, as shown in Table \ref{tab:Overall2}, emphasizing the advantages of fine-grained physical evolution over black-box models.

\subsection{PDE Kernel}
\label{sec:pdekernel}

We employ a set of five PDEs (\ref{equ:1}-\ref{equ:5}) including the motion
equation, the continuous equation and others to establish a closed system, which simulate the physical evolution of 5 essential atmospheric variables: $u$ (latitude-direction wind), $v$ (longitude-direction wind), $z$ (geopotential), $q$ (humidity), $T$ (temperature). The partial derivative of each atmospheric variable with respect to time can be separated mathematically (details in Appendix \ref{pdes}), denotes as $S_{PDE}$, which takes current weather state as input and produces the derivative of each variable with respect to time. We define PDE kernel $\mathcal{K}$ as the evolution of the variables over a short period of time $t_s$, as demonstrated in Equation \ref{equ:spde}.


\begin{equation}
\resizebox{11cm}{!}{$
S_{PDE}\left ( \mathcal X \right ) =\left\{\begin{matrix}
\frac{\partial u}{\partial t}=S_u(u,v,z,q,T) \ref{equ:10} \\[3pt]
\frac{\partial v}{\partial t}=S_v(u,v,z,q,T) \ref{equ:10} \\[3pt]
\frac{\partial z}{\partial t}=S_z(u,v,z,q,T) \ref{equ:15} \\[3pt]
\frac{\partial q}{\partial t}=S_q(u,v,z,q,T) \ref{equ:20} \\[3pt]
\frac{\partial T}{\partial t}=S_T(u,v,z,q,T) \ref{equ:12} \\[3pt]
\end{matrix}\right.,\ \ \ \ 
\begin{matrix}
\mathrm{PDE} \ \mathrm{Kernel} \ \  \mathcal{K}\left (  \mathcal X \right ) = S_{PDE}(\mathcal X) t_s+\mathcal X\\
\\
\mathrm{where} \ t_s=\frac{1}{12}t_{data} 
\end{matrix}
$}
\label{equ:spde}
\end{equation}


Calculating $S_{PDE}$ requires the use of differential and integral operations. For example, for temperature $T$, its derivative with respect to time is shown in Equation~\ref{equ:pde_example}. In order to efficiently calculate $S_{PDE}$ and enable loss backward~\cite{imambi2021pytorch}, we designed a fast implementation of differentiation and integration through convolution and matrix multiplication respectively. Equation~\ref{equ:conv_diff} presents the implementation of the differential and integral of $\mathcal X$ in the $x$ direction (latitude direction). 


\begin{equation}
\begin{aligned}
\frac{\partial T}{\partial t} = \frac{-L\frac{\partial z }{\partial p} w-\frac{\partial z }{\partial p}w  }{c_{p}}- u \frac{\partial T}{\partial x} - v \frac{\partial T}{\partial y} - w \frac{\partial T}{\partial p},\  \mathrm{where} \  w =-\int \left (\frac{\partial u }{\partial x} +\frac{\partial v }{\partial y}  \right ) \mathrm{d}p
\end{aligned}
\label{equ:pde_example}
\end{equation}


\begin{equation}
\resizebox{12cm}{!}{$
\left\{\begin{matrix}
\frac{\mathrm{d} \mathcal X}{\mathrm{d} x} = \frac{1}{12} Conv\left ( \mathcal X, K_x \right )\\
\\
\int \mathcal X \mathrm{d} x= \mathcal X M_x
\end{matrix}\right.,
K_x=\begin{bmatrix}
0 &0 &0 &0 &0\\
0 &0 &0 &0 &0\\
1 &-8 &0 &8 &-1\\
0 &0 &0 &0 &0\\
0 &0 &0 &0 &0
\end{bmatrix},
M_x=\begin{bmatrix}
1 &1 & \cdots &1 &1\\
0 &1 & \cdots &1 &1\\
\vdots & \vdots & \ddots &\vdots  & \vdots\\
0 &0 & \cdots &1 &1\\
0 &0 & \cdots  &0 &1
\end{bmatrix} \in \mathbb{R}^{W\times W}
$}
\label{equ:conv_diff}
\end{equation}

Similarly, $K_y$ and $M_y$ can be constructed to perform differential and integral operations in the $y$ direction (longitude direction). For differential and integral operations in the $p$ direction (pressure level direction), we first reshape $\mathcal X \in \mathbb{R}^{C \times H \times W}$ to 3D space $\mathcal X_{3D} \in \mathbb{R}^{\frac{C}{P} \times P \times H \times W}$ based on the variables' pressure layers, and then implement corresponding operations through $K_p$ and $M_p$.

\subsection{HybridBlock with Adaptive Router}

\begin{wrapfigure}[18]{r}{0.4\textwidth}
    \vspace{-13pt}
    \includegraphics[scale=0.56, trim={0 0 0 0}, clip]{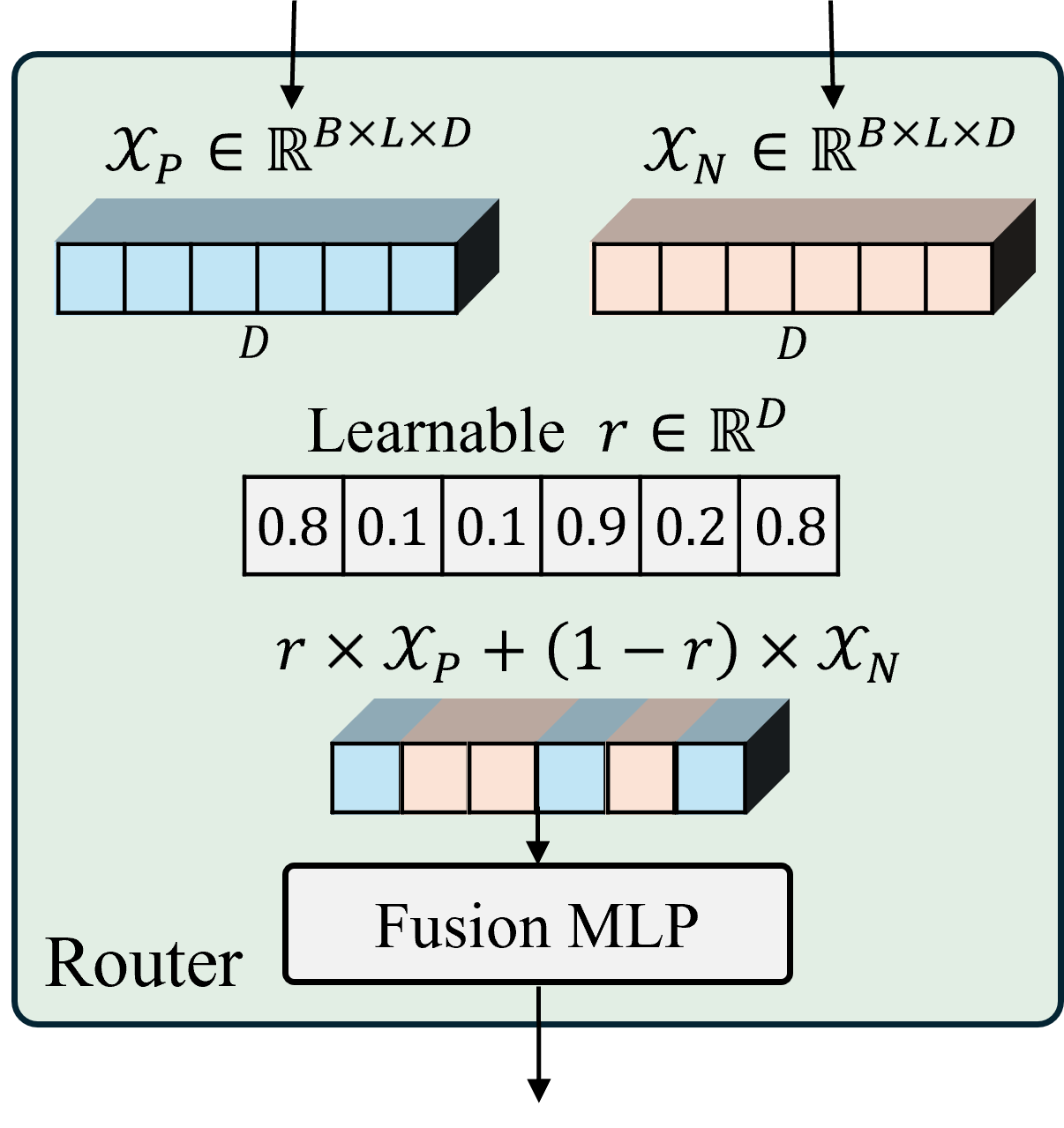}
    \caption{\textbf{Router in HybridBlock}. $B$ represents batch size, $L$ is the number of tokens with $D$ dimension.}
    \label{fig:router}
\end{wrapfigure}

HybridBlock is a module that combines physics and AI. Firstly, it employs neural networks to address the issue of error accumulation resulting from the stacking of PDE kernel $\mathcal{K}$. Secondly, it utilizes the PDE kernel $\mathcal{K}$ to guide the neural networks to learn the physical evolution of a specific time step. The structure of HybridBlock consists of three PDE kernels $\mathcal{K}$ and one parallel Attention Block. Consequently, the time step corresponding to a HybridBlock is $t_{block}=3\times t_s=\frac{1}{4}t_{data}$.

HybridBlock has two branches, as depicted in Figure~\ref{fig: Framework}, one is physics and the other is AI. The neural networks features $\mathcal{X}_{N}$ are aligned with physical features $\mathcal{X}_{P}$ through a convolutional layer, followed by three PDE kernels. Subsequently, the PDE kernel output is projected back to the latent space of $\mathcal{X}_{N}$ through another convolutional layer. Finally, features fusion is performed through the learnable router shown in Figure~\ref{fig:router}. 

In the router, the features $\mathcal{X}_{N}$ obtained from the neural networks and the features $\mathcal{X}_{P}$ derived from the PDE kernels are initially linearly fused along the feature dimension $D$, with the learnable factor $r$ initialized as $0.5:0.5$. Subsequently, the preliminary fused features will go through an Multilayer Perceptron~\cite{longstaff1987pattern} layer containing a ReLU~\cite{li2017convergence} activation function to accomplish nonlinear feature fusion.

\subsection{Lead Time Conditional Decoder}

HybridBlock provides the smallest time scale of model evolution, which is $t_{block}=\frac{1}{4}t_{data}$. Through $L\times$ HybridBlocks, we can predict the weather at a lead time of $\frac{L}{4}t_{data}$. To enable the model to generalize its prediction capabilities to finer-grained time scales, we design a lead time conditional decoder to generate forecasts varying lead times from the output of the corresponding HybridBlock. 

In order to promote the expression of the condition, we embed the lead time $t$ into a high-dimensional vector $t_{emb}$ through learnable Fourier embedding~\cite{salimans2022progressive}, as shown in Equation~\ref{equ:temb}.

\begin{equation}
t_{emb} = \mathrm{sin}(\pi \cdot t\cdot W)\oplus \mathrm{cos}(\pi \cdot t\cdot W) \oplus t, \ \mathrm{where}\  t\  \mathrm{is}\ \mathrm{lead}\ \mathrm{time} 
\label{equ:temb}
\end{equation}

where $W$ is a learnable vector of size 16, and $\oplus$ denotes concatenation. Furthermore, $t_{emb}$ will be concatenated with the output of HybridBlock and input to the decoder together. The decoder structure utilizes a Swin Transformer~\cite{liu2021swin} with hierarchical upsampling, as illustrated in Figure \ref{fig: Framework}.

\subsection{Multiple Lead Time Training}

For dataset like ERA5~\cite{hersbach2020era5} or WeatherBench~\cite{rasp2020weatherbench}, their time resolution is $t_{data}=1\mathrm{h}$. We set the time step of the PDE kernel to $t_s=\frac{1}{12}t_{data}=300\mathrm{s}$. Consequently, the time step of each HybridBlock is $t_{block}=3\times t_{s}=900\mathrm{s}$, equivalent to 15 minutes. By cascading 24 HybridBlocks, model can generate forecasts at a lead time of $24\times \text{15min} = 6\mathrm{h}$. To encourage the model to learn evolution for different lead times and generalize forecasting to finer-grained time scales, during training, we not only use the output of the last HybridBlock but also include the outputs of the 4th and 12th HybridBlocks. These outputs are passed through the lead time conditional decoder with corresponding $t_{emb}$ to predict the weather states at $4\times \text{15min} = 1\mathrm{h}$ and $12\times \text{15min} = 3\mathrm{h}$.

During inference, we can take the output of the second HybridBlock and pass it through the decoder with corresponding $t_{emb}$ to get $2\times \text{15min} = \text{30min}$ forecasts, which are not present in the dataset. In the Section~\ref{sec:now}, we provide a comprehensive demonstration showcasing the accuracy of these generalized prediction results for time scales smaller than the dataset's time resolution.

\section{Experiment}

Through the design of HybridBlock mixed with physics \& AI and the multi-lead time training method, our model is capable of simultaneously conducting short-term forecasting and long-term forecasting without additional finetuning~\cite{martinez2021permute} on different forecasting tasks. In the experiments, we will showcase the superior performance of our model and try to answer the following questions: 

\textbf{(1)} How does the model perform on the medium-range forecasting task?


\textbf{(2)} How does the model perform on the \textbf{\textit{generalized 30-minute nowcasting}} task?


\textbf{(3)} As a hybrid expert model of AI and physics, what roles do they each play?


\textbf{(4)} How do PDE kernel and multi-lead time training contribute to the overall performance?

\subsection{Experimental Setup}
\label{sec:ExperimentalSetup}

\begin{wraptable}[7]{r}{7cm}
\vspace{-1.5em}
\addtolength{\tabcolsep}{1.0pt}
\renewcommand{\arraystretch}{0.8}
    \centering
    \resizebox{7cm}{!}{
    \begin{tabular}{p{1.8cm}p{0.5cm}p{0.5cm}p{2.5cm}}
    \toprule
    Dataset         & Train      & Test        & Time resolution \\
    \midrule
    WeatherBench    & \multicolumn{1}{c}{\checkmark} & \multicolumn{1}{c}{\checkmark}  & \multicolumn{1}{c}{1-hour}\\
    NASA            & \multicolumn{1}{c}{\texttimes} & \multicolumn{1}{c}{\checkmark}  & \multicolumn{1}{c}{30-minute}\\
    \bottomrule
    \end{tabular}
    }
    \caption{\small{\textbf{Datasets.} NASA dataset only contains precipitation, which will be used as the ground truth for precipitation nowcast.}}\label{tab:dataset}
\end{wraptable}

\paragraph{Dataset.} We use WeatherBench~\cite{rasp2020weatherbench} as our training dataset, whose time resolution is $t_{data}=1\text{h}$ and spatial resolution is $128\times 256$. The dataset spanning from 1980 to 2015 serves as training set, while the data of 2017 is the validation and test sets. Our model processes 4 surface variables and 5 upper-air variables across 13 pressure levels, as shown in Table \ref{tab:variables}.

\begin{wraptable}[12]{r}{7cm}
\vspace{0.8em}
\addtolength{\tabcolsep}{1.0pt}
\renewcommand{\arraystretch}{0.8}
    \centering
    \vspace{-0.6em}
    \resizebox{7cm}{!}{
    \begin{tabular}{p{0.5cm}p{4.3cm}p{0.8cm}}
    \toprule
    Name & Description & Levels \\
    \midrule
    u10    & x-direction wind at 10m height  & Single\\
    v10    & y-direction wind at 10m height  & Single\\
    t2m    & Temperature at 2m height        & Single\\
    tp     & Hourly precipitation            & Single\\
    z      & Geopotential                    & 13\\
    q      & Specific humidity               & 13\\
    u      & x-direction wind                & 13\\
    v      & y-direction wind                & 13\\
    T      & Temperature                     & 13\\
    \bottomrule
    \end{tabular}
    }
    \caption{\small{\textbf{Atmospheric Variables Considered.} The 13 levels are 50, 100, 150, 200, 250, 300, 400, 500, 600, 700, 850, 925, 1000 hPa.}}\label{tab:variables}
\end{wraptable}


Given that WeatherBench lacks data at finer temporal resolutions, we use the 30-minute satellite observations downloaded from \href{https://disc.gsfc.nasa.gov/}{NASA} as ground truth to quantitatively assess the model's generalizability. NOTE: Data from NASA is only used for testing and not for model training.


\paragraph{Tasks.} We conducted experiments on two typical weather forecasting tasks: medium-range forecasting and precipitation nowcasting. The forecast range for medium-range forecasting spans from 6 hours to 5 days, while the nowcasting is set to a range of 30 minutes to 2 hours.


\paragraph{Baseline Methods.} We compare WeatherGFT with four forecast approaches: FourCastNet~\cite{kurth2023fourcastnet} uses AFNO~\cite{guibas2021adaptive} networks to simulate the nonlinear relationship between weather variables, Keisler~\cite{keisler2022forecasting} models global atmospheric data through GNN, ClimODE\cite{verma2024climode} adds ordinary differential equations (ODE)~\cite{ince1956ordinary} to the neural networks, and ECMWF-IFS~\cite{persson2007user} is a physical dynamic model.

The above three data-driven models cannot generalize forecasting to finer-grained time scales due to the absence of 30-minute labels. Therefore, in nowcasting tasks, we interpolate the 30-minute forecast results through SOTA frame interpolation models Flavr~\cite{kalluri2023flavr} and UPR~\cite{jin2023unified}. In contrast, our model can conduct 30-minute predictions inherently without interpolating.

\paragraph{Implementation Details.} We implemented the model with PyTorch~\cite{imambi2021pytorch} and trained 50 epochs on 8 NVIDIA A100 GPUs~\cite{choquette2020nvidia} for 3 days, with a learning rate of cosine schedule starting from 5e-4.

\subsection{Skillful Medium-Range Forecasts by WeatherGFT}

\label{sec:mid}

Autoregression is commonly employed in medium-term forecasting, where the model output serves as the input for the subsequent forecast step, allowing for longer lead time predictions. However, prediction errors tend to accumulate during the autoregression, leading to an increase in the root mean square error (RMSE). As a result, a smaller RMSE indicates a more accurate prediction.

\begin{figure}[h]
\centerline
{\includegraphics[width=15cm]{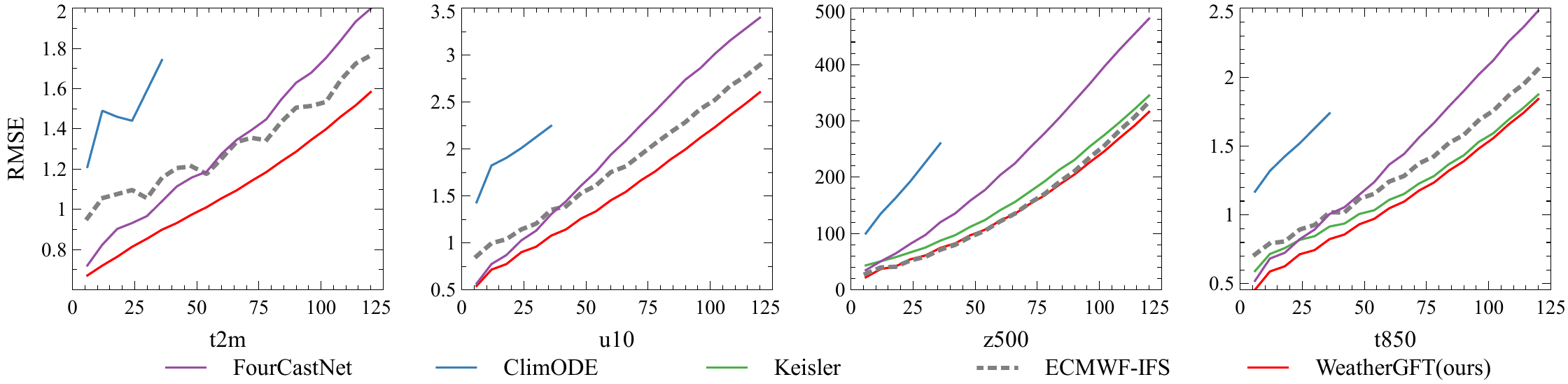}}
\caption{\textbf{Medium-Range Forecast.} The x-axis represents the lead time in hours, while the y-axis represents the RMSE for different variables. The smaller RMSE the better.}
\label{fig: rmse}
\end{figure}

\begin{figure}[t]
\centerline
{\includegraphics[width=15cm]{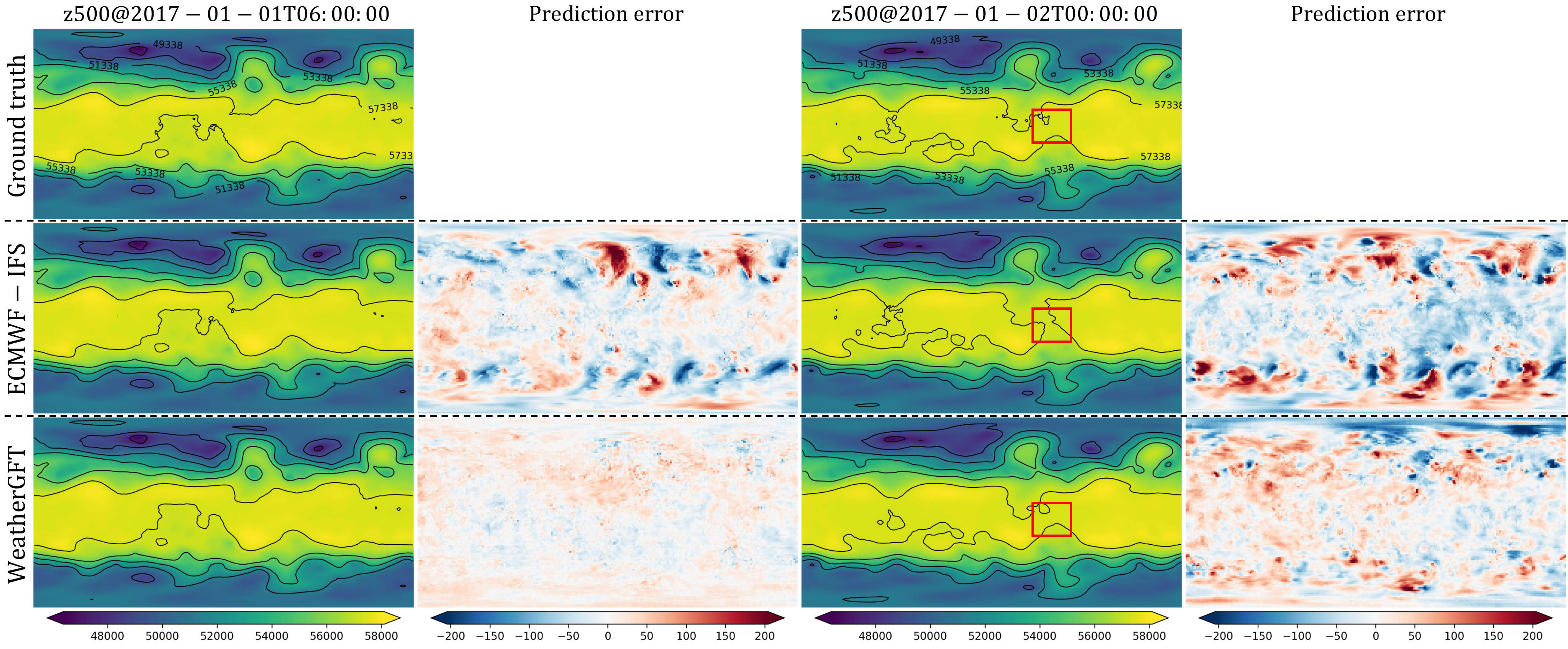}}
\caption{\textbf{Visualization of z500 Predictions.}}
\label{fig: compare}
\end{figure}

Figure \ref{fig: rmse} illustrates the changes in prediction RMSE of different weather variables as lead time increases. Our model demonstrates competitive performance across various lead times with AI or physical dynamics models, especially the prediction of surface temperature (t2m) and surface wind speed (u10) is significantly better than other models. The geopotential of the 500hpa pressure layer (z500) is a crucial weather variable in weather forecasting, as it reflects atmospheric circulation~\cite{oort1971atmospheric}, subtropical high-pressure systems~\cite{lashkari2019study}, and other significant phenomena. Due to the modeling of geopotential in the PDE \ref{equ:15}, z500 prediction of our model outperforms the physical dynamic model ECMWF-IFS as visualized in Figure \ref{fig: compare}.

From the visualization in Figure \ref{fig: compare}, our model is more accurate in predicting the subtropical high, as indicated by the highlighted red box. In addition, the prediction error of our model at the lead time of 6-hour is significantly smaller than that of the physical dynamic model ECMWF-IFS.

\subsection{Generalizing to Fine-grained Time Scale for Nowcasting}
\label{sec:now}

In contrast to conventional black-box AI models~\cite{kurth2023fourcastnet,keisler2022forecasting,xu2024extremecast} used in medium-range weather forecasting, WeatherGFT has the ability to break through the time scale limitations of the dataset, making the generalization to fine-grained temporal scales possible. This capability is facilitated by the dynamic progression of our PDE kernel modeling and multiple lead time training. Specifically, we use the second HybridBlock of the total 24 HybridBlocks to generate 30-minute generalized forecasts through the lead time conditional decoder, which is very important for precipitation nowcasting. 

To quantify the accuracy of the model's generalized nowcasting, we utilize the \href{https://disc.gsfc.nasa.gov/}{NASA} satellite precipitation observation dataset as the ground truth, which has a time resolution of 30-minute. We evaluate forecasts at 30, 60, 90, and 120 minutes. It is important to note that data of NASA were not used for training. For other comparison models that cannot directly produce half-hour forecasts, we use the frame interpolation models (i.e., Flavr~\cite{kalluri2023flavr} and UPR~\cite{jin2023unified}) to generate 30-minute predictions.

\begin{table*}[h]
    \centering
    \renewcommand{\arraystretch}{1.2}
    \resizebox{14cm}{!}{
        \begin{tabular}{lp{0.5cm}p{0.5cm}p{0.5cm}p{0.5cm}p{0.5cm}p{0.5cm}p{0.5cm}p{0.5cm}p{0.5cm}p{0.5cm}p{0.5cm}p{0.9cm}}
        \toprule
                   &  \multicolumn{3}{c|}{30-min}                   & \multicolumn{3}{c|}{60-min}                                       & \multicolumn{3}{c|}{90-min}                                       & \multicolumn{3}{c}{120-min}                  \\
                   [-5pt]
                  & {\fontsize{8}{1}\selectfont CSI$\uparrow$} & {\fontsize{8}{1}\selectfont CSI$\uparrow$} & \multicolumn{1}{c|}{{\fontsize{8}{1}\selectfont RMSE$\downarrow$}} & {\fontsize{8}{1}\selectfont CSI$\uparrow$} & {\fontsize{8}{1}\selectfont CSI$\uparrow$} & \multicolumn{1}{c|}{{\fontsize{8}{1}\selectfont RMSE$\downarrow$}} & {\fontsize{8}{1}\selectfont CSI$\uparrow$} & {\fontsize{8}{1}\selectfont CSI$\uparrow$} & \multicolumn{1}{c|}{{\fontsize{8}{1}\selectfont RMSE$\downarrow$}} & {\fontsize{8}{1}\selectfont CSI$\uparrow$} & {\fontsize{8}{1}\selectfont CSI$\uparrow$} & \multicolumn{1}{c}{{\fontsize{8}{1}\selectfont RMSE$\downarrow$}} \\ 
                  [-5pt]
                  & {\fontsize{8}{0.5}\selectfont @0.5} & {\fontsize{8}{0.5}\selectfont @1.5} & \multicolumn{1}{c|}{{\fontsize{8}{0.5}\selectfont tp1h}} & {\fontsize{8}{0.5}\selectfont @0.5} & {\fontsize{8}{0.5}\selectfont @1.5} & \multicolumn{1}{c|}{{\fontsize{8}{0.5}\selectfont tp1h}} & {\fontsize{8}{0.5}\selectfont @0.5} & {\fontsize{8}{0.5}\selectfont @1.5} & \multicolumn{1}{c|}{{\fontsize{8}{0.5}\selectfont tp1h}} & {\fontsize{8}{0.5}\selectfont @0.5} & {\fontsize{8}{0.5}\selectfont @1.5} & \multicolumn{1}{c}{{\fontsize{8}{0.5}\selectfont tp1h}} \\ 
\cline{2-13} 
FourCast+\textcolor{gray!90}{\textbf{Flavr}}     & \cellcolor{gray!20} 0.26 & \cellcolor{gray!20} 0.09 & \multicolumn{1}{c|}{\cellcolor{gray!20} 0.67} & 0.61 & 0.49 & \multicolumn{1}{c|}{0.24} & \cellcolor{gray!20} 0.25 & \cellcolor{gray!20} 0.09 & \multicolumn{1}{c|}{\textbf{\cellcolor{gray!20} 0.65}} & 0.37 & 0.26 & 0.46\\
FourCast+\textcolor{gray!90}{\textbf{UPR}}     & \cellcolor{gray!20} 0.20 &\cellcolor{gray!20}  0.10 & \multicolumn{1}{c|}{\cellcolor{gray!20} 0.76} & 0.61 & 0.49 & \multicolumn{1}{c|}{0.24} & \cellcolor{gray!20} 0.11 & \cellcolor{gray!20} 0.05 & \multicolumn{1}{c|}{\cellcolor{gray!20} 1.49} & 0.37 & 0.26 & 0.46\\
Keisler+\textcolor{gray!90}{\textbf{Flavr}}     & \cellcolor{gray!20} 0.25 & \cellcolor{gray!20} 0.09 & \multicolumn{1}{c|}{\textbf{\cellcolor{gray!20} 0.66}} & 0.59 & 0.48 & \multicolumn{1}{c|}{0.23} & \cellcolor{gray!20} 0.25 & \cellcolor{gray!20} 0.08 & \multicolumn{1}{c|}{\cellcolor{gray!20} 0.66} & 0.41 & 0.29 & 0.35\\
Keisler+\textcolor{gray!90}{\textbf{UPR}}      & \cellcolor{gray!20} 0.26 & \cellcolor{gray!20} 0.13 & \multicolumn{1}{c|}{\cellcolor{gray!20} 0.69} & 0.59 & 0.48 & \multicolumn{1}{c|}{0.23} & \cellcolor{gray!20} 0.26 & \cellcolor{gray!20} 0.13 & \multicolumn{1}{c|}{\cellcolor{gray!20} 0.68} & 0.41 & 0.29 & 0.35\\
ClimODE+\textcolor{gray!90}{\textbf{Flavr}}     &\cellcolor{gray!20} 0.26 & \cellcolor{gray!20} 0.09 & \multicolumn{1}{c|}{\cellcolor{gray!20} 0.67} & 0.62 & \textbf{0.51} & \multicolumn{1}{c|}{0.22} & \cellcolor{gray!20} 0.25 & \cellcolor{gray!20} 0.09 & \multicolumn{1}{c|}{\cellcolor{gray!20} 0.66} & 0.47 & 0.34 & 0.32\\
ClimODE+\textcolor{gray!90}{\textbf{UPR}}      & \cellcolor{gray!20} 0.25 & \cellcolor{gray!20} 0.12 & \multicolumn{1}{c|}{\cellcolor{gray!20} 0.67} & 0.62 & 0.49 & \multicolumn{1}{c|}{0.21} & \cellcolor{gray!20} 0.25 & \cellcolor{gray!20} 0.11 & \multicolumn{1}{c|}{\cellcolor{gray!20} 0.66} & 0.46 & 0.32 & 0.31\\
\rowcolor{blue!15} WeatherGFT(ours)      &  \textbf{0.28} & \textbf{0.17} & \multicolumn{1}{c|}{0.72} & \textbf{0.62} & 0.50 & \multicolumn{1}{c|}{\textbf{0.21}} & \textbf{0.28} & \textbf{0.16} & \multicolumn{1}{c|}{0.71} & \textbf{0.54} & \textbf{0.40} & \textbf{0.27}\\

        \bottomrule
        \end{tabular}
    }
    \caption{\textbf{Generalized Nowcast.} 60-min and 120-min are trained lead times, while 30-min and 90-min are generalized lead times. \textcolor{gray!90}{\textbf{Gray}} represents the results obtained through the frame interpolation model, \textcolor{blue!40}{\textbf{purple}} indicates the results obtained through our unified model without interpolating. For precipitation nowcasting, CSI (Critical Success Index) is the most important metric.}
    \label{tab:Overall2}
\end{table*}

\begin{figure}[t]
\centerline
{\includegraphics[width=14.5cm]{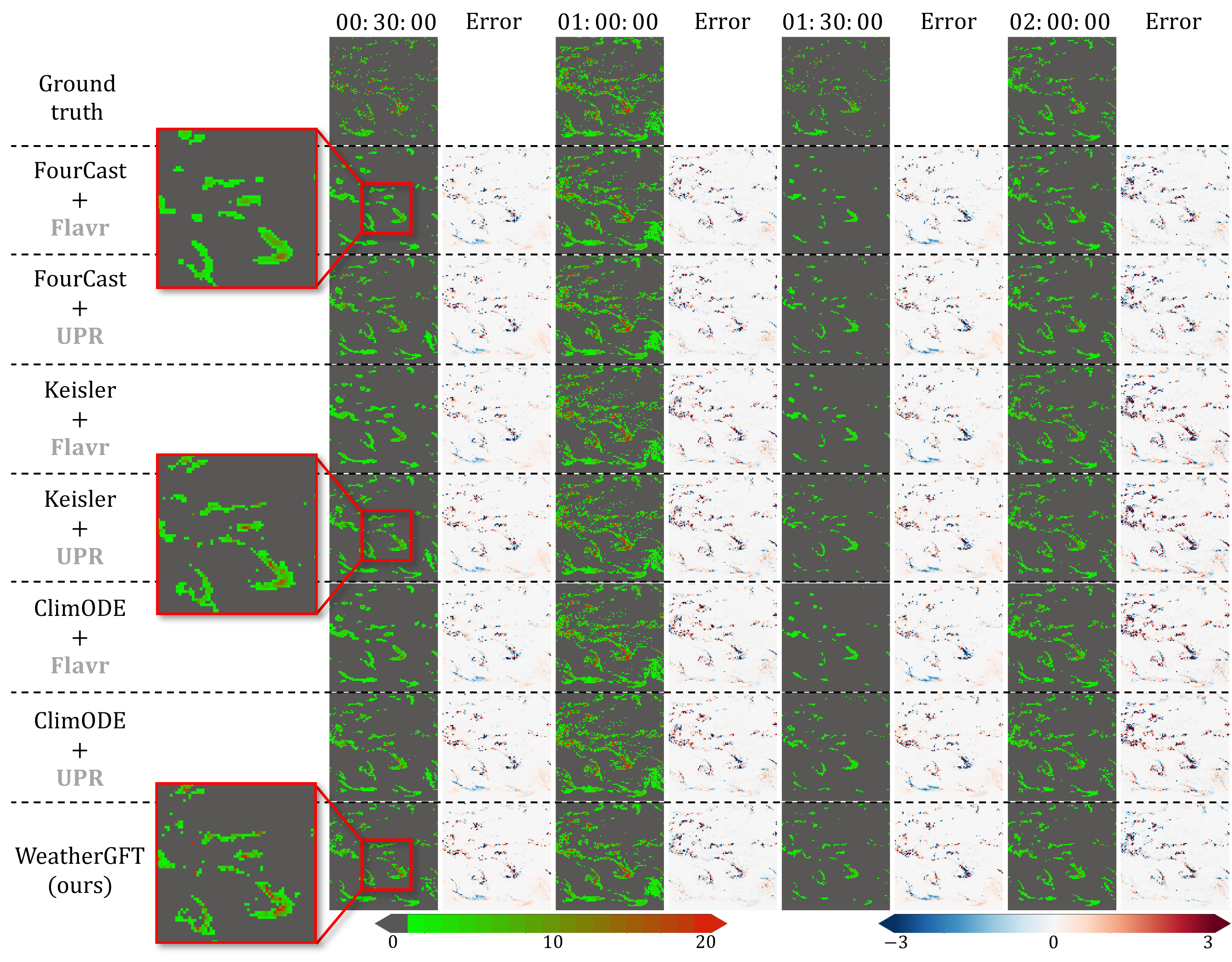}}
\caption{\textbf{Visualization of Precipitation Nowcast.} Precipitation in the area ranging from 34N to 50S and 148E to 128W during the time period from 00:00 to 02:00 on July 1, 2017.}
\label{fig: tp}
\end{figure}

$\mathrm{CSI@th}$ (Critical Success Index) refers to the hit rate of the area that reaches the threshold precipitation value $\mathrm{th}$. $\mathrm{CSI@0.5}$ can reflect the overall forecast accuracy in rainy areas, and $\mathrm{CSI@1.5}$ reflects the forecast accuracy in moderate rainy areas. Table \ref{tab:Overall2} shows that our model surpasses others across different lead times, especially in forecasting regions of moderate rainfall, i.e., $\mathrm{CSI@1.5}$.

The visualization in Figure \ref{fig: tp} reveals that when using frame interpolation to obtain 30-minute predictions, there is blurring occurring at different scales, resulting in the loss of extreme values, as indicated in the red box. Our model, which incorporates physical constraints, provides clearer predictions retaining extreme values without the need for frame interpolation.

\subsection{Weather Forecasts can Benefit from Physics and AI via WeatherGFT}
\label{sec:weight}

\begin{figure}[h]
\centerline
{\includegraphics[width=14.5cm]{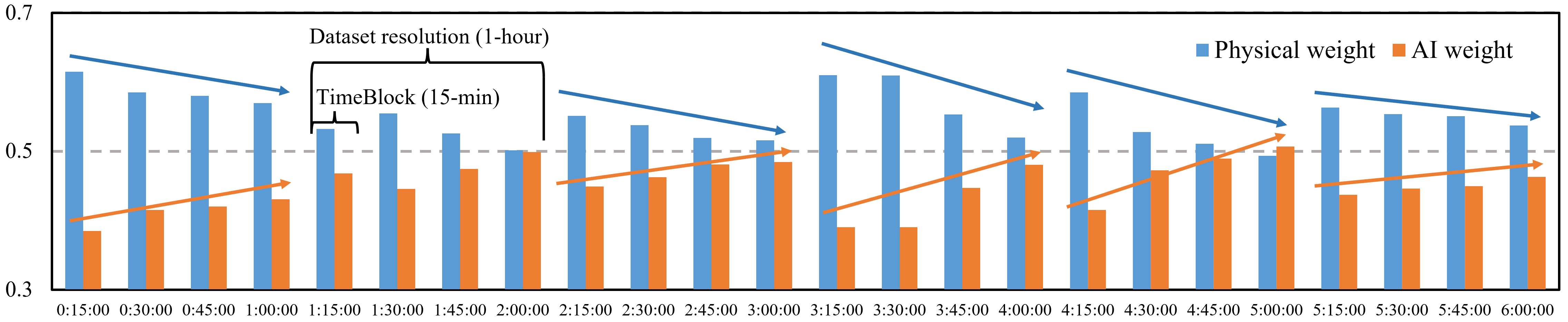}}
\caption{\textbf{The Weights in the Router of 24 HybridBlocks.}}
\label{fig: ratio}
\end{figure}

As a hybrid model combining both physics and AI components, it is crucial to analyze their contributions to the prediction process. We present insights into their respective proportions by visualizing the weight parameter $r$ within the learnable router (refer to Figure \ref{fig:router}). The visualization in Figure \ref{fig: ratio} reveals that the weights of the 24 HybridBlocks display a similar distribution:

\textbf{a)} The physical weight of the vast majority of HybridBlocks is significantly higher than the weight of AI, which shows that in the process of simulating time evolution, the PDE kernel plays a more important role, while the Attention Block only plays a supportive correction role. 
\textbf{b)} The physical weight gradually decreases while the weight of AI increases throughout each hour (dataset time resolution). This aligns with our underlying motivation, which acknowledges that errors may accumulate over time in the physics-based evolution. Consequently, a greater emphasis on AI corrections becomes necessary to compensate for these accumulated errors.

By averaging the $4\times 6$ HybridBlocks into $4$ time steps, the average weight every 15-minute is obtained in Figure \ref{fig:ratio_mean}, which shows the above two conclusions more clearly. To summarize, physics plays the main evolutionary role in the model, while AI plays an dynamic corrective role.

\subsection{Ablation Studies}

\begin{wraptable}[8]{r}{7cm}
\vspace{-0.6em}
\addtolength{\tabcolsep}{1.0pt}
\renewcommand{\arraystretch}{1.2}
    \centering
    \vspace{-0.6em}
    \resizebox{7cm}{!}{
    \begin{tabular}{p{2.2cm}p{0.2cm}p{0.3cm}p{0.3cm}|p{0.3cm}p{0.3cm}|p{0.3cm}p{0.3cm}}
    \toprule
                     & \multicolumn{1}{c|}{30-min} & \multicolumn{2}{c|}{RMSE@1-h} & \multicolumn{2}{c|}{RMSE@6-h} & \multicolumn{2}{c}{RMSE@3-d} \\
                     & \multicolumn{1}{c|}{nowcast} & t2m$\downarrow$            & z500$\downarrow$            & t2m$\downarrow$            & z500$\downarrow$            & t2m$\downarrow$           & z500$\downarrow$           \\ \cline{2-8} 
    Attent Block  & \multicolumn{1}{c|}{\texttimes} & 0.52           & 18.76           & 0.73           & 24.21           & 1.23          & 157.9          \\
    \ \ + PDE Kernel     & \multicolumn{1}{c|}{\checkmark} & 0.57           & 20.43           & 0.70            & \textbf{21.78}           & 1.22          & 153.8           \\
    \ \ \ \ \ + Muti Time & \multicolumn{1}{c|}{\checkmark} & \textbf{0.49}           & \textbf{16.66}           & \textbf{0.67}           & 21.80           & \textbf{1.14}          & \textbf{152.4}          \\
    \bottomrule
    \end{tabular}
    }
    \caption{\small{\textbf{Ablation Experiment.}}}\label{tab:Ablation}
\end{wraptable}

We use Swin Attention Block~\cite{liu2021swin} as the baseline for the ablation studies. For this baseline networks without PDE kernel constraints, as a black-box model, it will only learn the mapping of data pairs corresponding to the lead time. Consequently, its internal information between blocks is unexplainable, which also results in being unable to predict moments without data labels, such as 30-minute nowcasting.

\begin{wrapfigure}[12]{r}{7cm}
    \vspace{-55pt}
    \includegraphics[scale=0.5, trim={0 0 0 0}, clip]{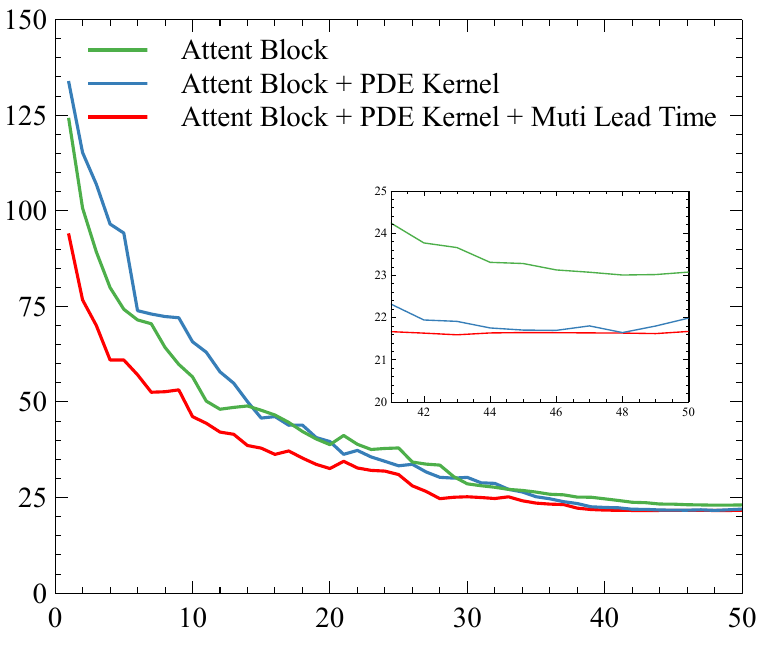}
    \caption{\textbf{RMSE z500 as Training Epochs.}}
    \label{fig:z500merge}
\end{wrapfigure}

\textbf{PDE kernel is crucial to the generalization of finer-grained predictions.} Instead of simply learning the mapping between data, the model learns the evolution of the corresponding time step according to the physics laws, making information of each neural network layer explainable, thereby facilitating generalized 30-minute nowcasting. In addition, we find that the introduction of the PDE kernel also improved the prediction accuracy of the model. 

\textbf{Multiple lead time training accelerates convergence and improves the accuracy of model prediction}, as shown in Figure \ref{fig:z500merge}. We hypothesize that this phenomenon can be attributed to the loss backward from different lead times, which alleviates the issue of vanishing gradients~\cite{hochreiter1998vanishing}, allowing the parameters of different layers to quickly warm up and improve the expression of the model.

\section{Conclusion}
\label{sec: Conclusion}

Most existing data-driven weather forecast methods which operated as black-box models via purely performing data mapping are unable to generalize at finer temporal scale beyond the inherent time resolution of the training datasets due to the absence of the fine-grained physics modeling. This paper proposes a physics-AI hybrid model to solve this problem. Through the exquisitely designed PDE kernel, each block in the networks can simulate the evolution of physical variables at finer-gained time step, while AI plays the role of adaptive correction, which makes our model capable of generalizing predictions to a finer time scale beyond dataset. By employing our proposed multi-lead time training strategy, our model trained on an hourly dataset exhibits remarkable ability of generalized 30-minute forecasts, while maintaining prediction errors that are competitive with those of pure AI and physical models in both medium-range forecast and precipitation nowcast.

The main limitation of our model is that only five important atmospheric equations are currently considered, which is still far from fully modeling the atmospheric motion process. Another limitation of this paper is that the experiments have been conducted solely at a spatial resolution of $128 \times 256$. As part of our future work, we plan to extend our experiments to higher resolutions such as $721 \times 1440$ to assess the model's performance under different settings. Additionally, while the minimum evolution time scale of our model is 15 minutes, we were unable to evaluate 15-minute generalized predictions due to the absence of corresponding validation data at that specific time scale. Therefore, we are currently only able to perform evaluations of 30-minute generalized predictions.

For future work, we plan to incorporate additional physical laws into our model and conduct higher-resolution experiments to ascertain the upper limit of its capabilities.

\clearpage

\section*{Acknowledgements}

This work is supported by Shanghai Artificial Intelligence Laboratory.

\bibliographystyle{plain}
\bibliography{references.bib}


\clearpage
\appendix

\section{PDE Solver}
\label{pdes}

We constrain 5 atmospheric variables, that is, $u$ (latitude-direction wind), $v$ (longitude-direction wind), $z$ or 
 $\phi$ (geopotential), $q$ (humidity), $T$ (temperature), through the following set of five partial differential equations (PDEs)~\cite{kimura2002numerical}:

\begin{equation}
\frac{\mathrm{d} \textbf{V}  }{\mathrm{d} t}+f \textbf{k}\times \textbf{V} = -g\nabla _{p}z+\textbf{F}_{h}
\label{equ:1}
\end{equation}

\begin{equation}
\frac{\partial \phi }{\partial p} = -\frac{1}{\rho }  
\label{equ:2}
\end{equation}

\begin{equation}
\nabla _{p}\cdot \textbf{V}  + \frac{\partial w}{\partial p} = 0 
\label{equ:3}
\end{equation}

\begin{equation}
c_{p}\frac{\mathrm{d} T}{\mathrm{d} t} -\frac{1}{\rho }w = Q
\label{equ:4}
\end{equation}

\begin{equation}
p=\rho RT
\label{equ:5}
\end{equation}

The expansion of $\frac{\mathrm{d} }{\mathrm{d} t}$  is as follows:

\begin{equation}
\frac{\mathrm{d} }{\mathrm{d} t} =\left ( \frac{\partial }{\partial t}  \right )_{p} +\textbf{V}\cdot \nabla _{p}\left ( \  \right ) +w\frac{\partial }{\partial p}  
\label{equ:6}
\end{equation}

The PDE above is in the pressure coordinate system, which is aligned with the input to our model, as the input to the model comes from 13 pressure layers. In the air pressure coordinate system, the following equation is also satisfied:

\begin{equation}
\frac{\partial p}{\partial t} = 0
\label{equ:7}
\end{equation}

$w$ represents the vertical wind speed and is not directly included as one of the input variables in our model. However, it can be derived from $u$ and $v$ using following equation:

\begin{equation}
\begin{aligned}
\frac{\partial w}{\partial p}  &= -\frac{\partial u }{\partial x} -\frac{\partial v }{\partial y}\\
w &=-\int \left (\frac{\partial u }{\partial x} +\frac{\partial v }{\partial y}  \right ) \mathrm{d}p
\end{aligned}
\label{equ:8}
\end{equation}

After getting $w$, we can get $\frac{\partial u}{\partial t}$ and $\frac{\partial v}{\partial t}$ according to Equation \ref{equ:1}.

\begin{equation}
\left\{\begin{matrix}
\begin{aligned}
\frac{\partial u}{\partial t} + u\frac{\partial u}{\partial x}+ v\frac{\partial u}{\partial y} +w\frac{\partial u}{\partial p} - fv=-\frac{\partial \phi }{\partial x}  \\
\frac{\partial v}{\partial t} + u\frac{\partial v}{\partial x}+ v\frac{\partial v}{\partial y} +w\frac{\partial v}{\partial p} + fu=-\frac{\partial \phi }{\partial y}
\end{aligned}
\end{matrix}\right.
\label{equ:9}
\end{equation}

\begin{equation}
\left\{\begin{matrix}
\begin{aligned}
\frac{\partial u}{\partial t} =- u\frac{\partial u}{\partial x}- v\frac{\partial u}{\partial y} -w\frac{\partial u}{\partial p} + fv-\frac{\partial \phi }{\partial x}  \\
\frac{\partial v}{\partial t} =- u\frac{\partial v}{\partial x}- v\frac{\partial v}{\partial y} -w\frac{\partial v}{\partial p} - fu-\frac{\partial \phi }{\partial y}
\end{aligned}
\end{matrix}\right.
\label{equ:10}
\end{equation}

where $f=7.29e-5$ is a constant.

According to Equation \ref{equ:4}, we can get $\frac{\partial T}{\partial t}$:

\begin{equation}
\left\{\begin{matrix}
\begin{aligned}
&c_{p}\left ( \frac{\partial T}{\partial t} + u \frac{\partial T}{\partial x} +v \frac{\partial T}{\partial y} + w \frac{\partial T}{\partial p}  \right )-\frac{1}{\rho }w = Q  \\
&Q=-L\frac{\partial \phi }{\partial p} w
\end{aligned}
\end{matrix}\right.
\label{equ:11}
\end{equation}

\begin{equation}
\frac{\partial T}{\partial t} = \frac{-L\frac{\partial \phi }{\partial p} w-\frac{\partial \phi }{\partial p}w  }{c_{p}}- u \frac{\partial T}{\partial x} - v \frac{\partial T}{\partial y} - w \frac{\partial T}{\partial p}
\label{equ:12}
\end{equation}

where $L=2.5e6$ and $c_p = 1005$ are constants.

According to Equations \ref{equ:2} and Equations \ref{equ:5}, we can get $\frac{\partial \phi }{\partial t}$:

\begin{equation}
\frac{\partial \phi }{\partial p} = -\frac{1}{\rho } =-\frac{RT}{p} 
\label{equ:13}
\end{equation}

\begin{equation}
\begin{aligned}
\frac{\partial ^{2}\phi }{\partial p \partial t} &=-\frac{\partial \frac{RT}{p} }{\partial t} \\
&=-R\left ( \frac{1}{p}\frac{\partial T}{\partial t} -\frac{T}{p^2} \frac{\partial p}{\partial t}    \right ) \\
&=-\frac{R}{p} \frac{\partial T}{\partial t}
\end{aligned}
\label{equ:14}
\end{equation}

\begin{equation}
\begin{aligned}
\frac{\partial \phi }{\partial t}  &= \int \frac{\partial ^{2}\phi }{\partial p \partial t}\mathrm{d}p\\
&=-\int \frac{R}{p} \frac{\partial T}{\partial t}  \mathrm{d}p
\end{aligned}
\label{equ:15}
\end{equation}

where $R = 8.314$ is a constant.




Finally, according to the water vapor equation \ref{equ:19}, we can get $\frac{\partial q}{\partial t}$:

\begin{equation}
\left\{\begin{matrix}
\begin{aligned}
 &\frac{\mathrm{d} q}{\mathrm{d} t} = \frac{\delta F}{RT } \frac{\mathrm{d} \phi }{\mathrm{d} t}  \\
 &\delta = \left\{\begin{matrix}
\begin{aligned}
&0,\frac{\mathrm{d} \phi }{\mathrm{d} t}  < 0 \ and \ q\ge q_{s}\\
&1,else
\end{aligned}
\end{matrix}\right.\\
&F=q_{s}T\frac{LR-c_{p}R_vT}{c_pR_vT^2+L^2q_s}  \\
&e_s = 6.112\times exp\left ( \frac{17.67T'}{T'+243.5}  \right )  \\
&T' =  T - 273.15 \\
& q_s = \frac{0.622e_s}{p -0.378e_s} 
\end{aligned}
\end{matrix}\right.
\label{equ:19}
\end{equation}

\begin{equation}
\frac{\partial q}{\partial t} = \frac{\delta F}{RT }\left ( \frac{\partial \phi }{\partial t} +u\frac{\partial \phi }{\partial x}+v\frac{\partial \phi }{\partial y} +w\frac{\partial \phi }{\partial z}\right )-u\frac{\partial q}{\partial x}-v\frac{\partial q }{\partial y} -w\frac{\partial q }{\partial z} 
\label{equ:20}
\end{equation}

where $R_v = 461.5$ and $R_d = 287$ are constants.

\newpage
\section{Implementation of Integrals and Differentials}
\label{sec:code}

Integral in p-direction (pressure levels direction) is implemented with PyTorch~\cite{imambi2021pytorch} as follows:

\begin{python}
def integral_z(input_tensor):
    # Pressure-direction integral
    B, pressure_level_num, H, W = input_tensor.shape
    input_tensor = input_tensor.reshape(B, pressure_level_num, H*W)
    output = M_z.to(input_tensor.dtype).to(input_tensor.device) @ input_tensor
    output = output.reshape(B, pressure_level_num, H, W)
    return output
\end{python}

$M_x$ obtains the integral through matrix multiplication. Given the input matrix $x$ below, the result of $xM_x$ is:

\begin{equation}
x=\begin{bmatrix}
1 & 4\\
2 & 5\\
3 & 6
\end{bmatrix},\ xM_x=
\begin{bmatrix}
1 & 4\\
2 & 5\\
3 & 6
\end{bmatrix}
\begin{bmatrix}
1 & 1\\
0 & 1
\end{bmatrix}=\begin{bmatrix}
1 & 1+4\\
2 & 2+5\\
3 & 3+6
\end{bmatrix}.
\end{equation}

Differentials in x-direction (latitude direction) is implemented with PyTorch as follows:

\begin{python}
def d_x(input_tensor):
    # Latitude-direction differential
    B, C, H, W = input_tensor.shape
    conv_kernel = torch.zeros([1,1,1,5], device=input_tensor.device, dtype=input_tensor.dtype, requires_grad=False)
    conv_kernel[0,0,0,0] = 1
    conv_kernel[0,0,0,1] = -8
    conv_kernel[0,0,0,3] = 8
    conv_kernel[0,0,0,4] = -1

    input_tensor = torch.cat((input_tensor[:,:,:,-2:], 
                              input_tensor,
                              input_tensor[:,:,:,:2]), dim=3)
    _, _, H_, W_ = input_tensor.shape
    
    input_tensor = input_tensor.reshape(B*C, 1, H_, W_)
    output_x = F.conv2d(input_tensor, conv_kernel)/12
    output_x = output_x.reshape(B, C, H, W)
    output_x = output_x/pixel_x.to(output_x.dtype).to(output_x.device)
    
    return output_x
\end{python}

$K_x$ is the convolution kernel. Assume a one-dimensional input data $x=[-2,-1,0,1,2]$. It gradually increases from left to right by 1, that is, its gradient is 1. Applying convolution kernel $K_x$ to $x$, the result is: $Conv(x, K_x)=\frac{(-2)\times 1+(-1)\times (-8)+0\times 0+1\times 8+2\times (-1)}{12} = 1$.
By using this convolution kernel, the data gradient can be determined.

Differentials in y-direction (longitude direction) is implemented with PyTorch as follows:

\begin{python}
def d_y(input_tensor):
    # Longitude-direction differential
    B, C, H, W = input_tensor.shape
    conv_kernel = torch.zeros([1,1,5,1], device=input_tensor.device, dtype=input_tensor.dtype, requires_grad=False)
    conv_kernel[0,0,0] = -1
    conv_kernel[0,0,1] = 8
    conv_kernel[0,0,3] = -8
    conv_kernel[0,0,4] = 1

    input_tensor = torch.cat((input_tensor[:,:,:2], 
                              input_tensor,
                              input_tensor[:,:,-2:]), dim=2)
    _, _, H_, W_ = input_tensor.shape
    
    input_tensor = input_tensor.reshape(B*C, 1, H_, W_)
    output_y = F.conv2d(input_tensor, conv_kernel)/12
    output_y = output_y.reshape(B, C, H, W)
    output_y = output_y/pixel_y
    
    return output_y
\end{python}

Differentials in p-direction (pressure levels  direction) is implemented with PyTorch as follows:

\begin{python}
def d_z(input_tensor):
    # Pressure-direction differential
    conv_kernel = torch.zeros([1,1,5,1,1], device=input_tensor.device, dtype=input_tensor.dtype, requires_grad=False)
    conv_kernel[0,0,0] = -1
    conv_kernel[0,0,1] = 8
    conv_kernel[0,0,3] = -8
    conv_kernel[0,0,4] = 1

    input_tensor = torch.cat((input_tensor[:,:2], 
                              input_tensor,
                              input_tensor[:,-2:]), dim=1)
    
    input_tensor = input_tensor.unsqueeze(1) # B, 1, C, H, W
    output_z = F.conv3d(input_tensor, conv_kernel)/12
    output_z = output_z.squeeze(1)
    output_z = output_z/pixel_z.to(output_z.dtype).to(output_z.device)
    
    return output_z
\end{python}

\section{Hyperparameter Details}

\begin{table}[h]
    \centering
    \begin{tabular}{@{}ll@{}}
        \toprule
        Hyperparameter                & Value                   \\ 
        \midrule
        Max epoch                     & 50                      \\
        Batch size                    & 4x8 (GPUs)             \\
        Learning rate                 & 5e-4                    \\
        Learning rate schedule         & Cosine                  \\
        Patch size                    & 4x4                     \\
        Embedding dimension           & 1024                    \\
        MLP ratio                     & 4                       \\
        Activation function           & GLUE                    \\
        Input (0-hour)               & [4, 69, 128, 256]      \\
        Output (1, 3, 6-hour)        & [4, 3, 69, 128, 256]   \\ 
        \bottomrule
    \end{tabular}
    \caption{\textbf{Hyperparameters of the Model}}
    \label{tab:hyperparameters}
\end{table}

\begin{table}[h]
    \centering
    \resizebox{14cm}{!}{
    \begin{tabular}{@{}llllll@{}}
        \toprule
        Datasets      & Training set & Validation set & Test set     & Time resolution & Variable                        \\ \midrule
        WeatherBench  & 1980-2014    & 2015           & 2017-2018    & 1h              & tp, t2m, u10, v10, z, q, u, v, t \\
        NASA          & None         & None           & 2017-2018    & 30min           & tp                             \\ \bottomrule
    \end{tabular}
    }
    \caption{\textbf{Datasets Information}}
    \label{tab:datasets}
\end{table}

\section{Additional Experiments}

\subsection{Prediction Bias Evaluation}

Bias~\cite{ben2024rise, chen2023towards, xiaotowards} indicates the disparity between the model's predictions and the ground truth. Negative bias indicates underestimation, a prevalent issue in forecasting models. Although the PDE kernel was not specifically designed to address bias underestimation, experimental results indicate that its usage helps ameliorate underestimation.

\begin{equation}
\mathrm {bias} = \overline{pred-gt} 
\end{equation}

\begin{figure*}[!ht]
    \centering
    \includegraphics[width=4.5cm]{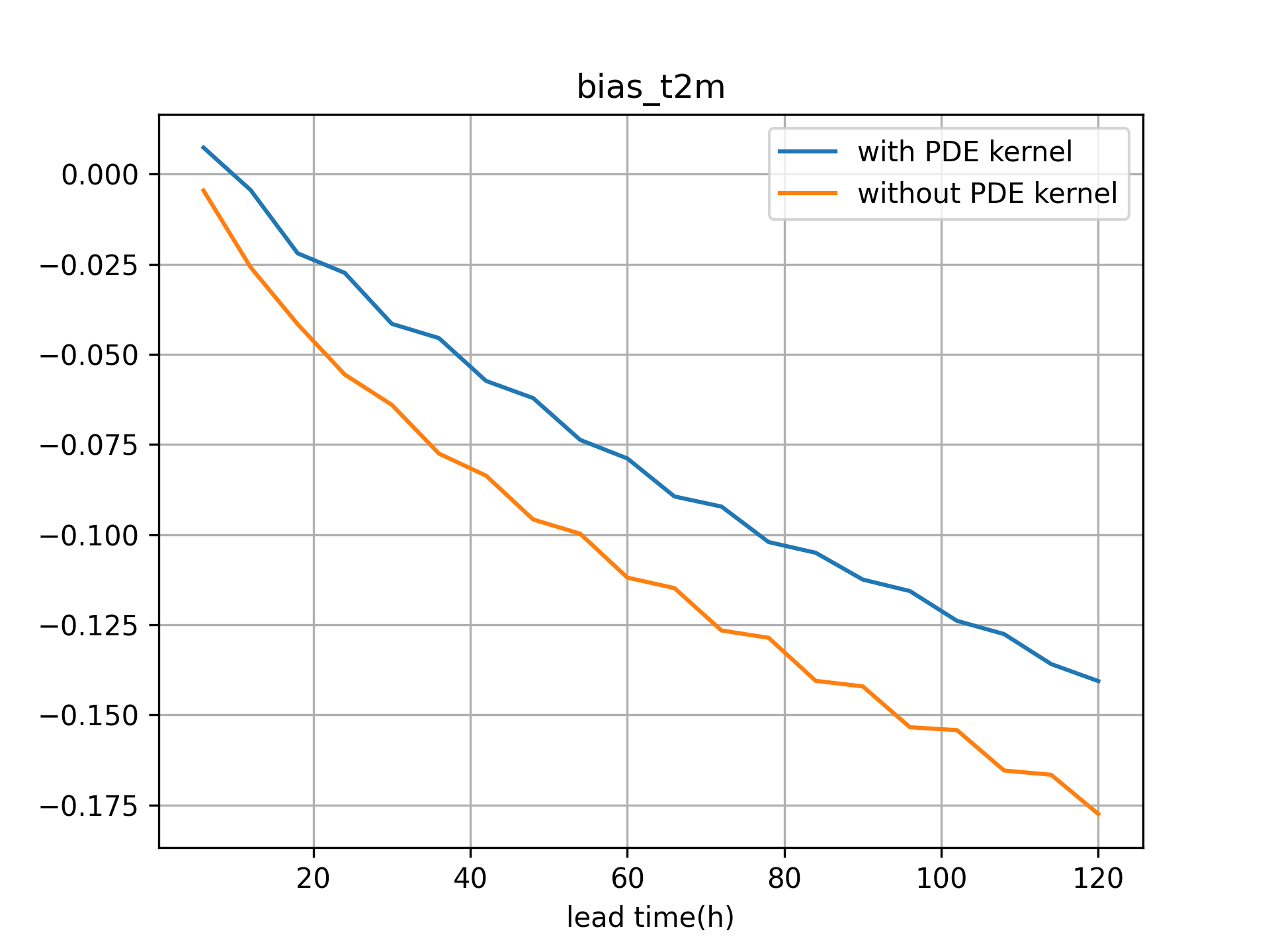} 
    \includegraphics[width=4.5cm]{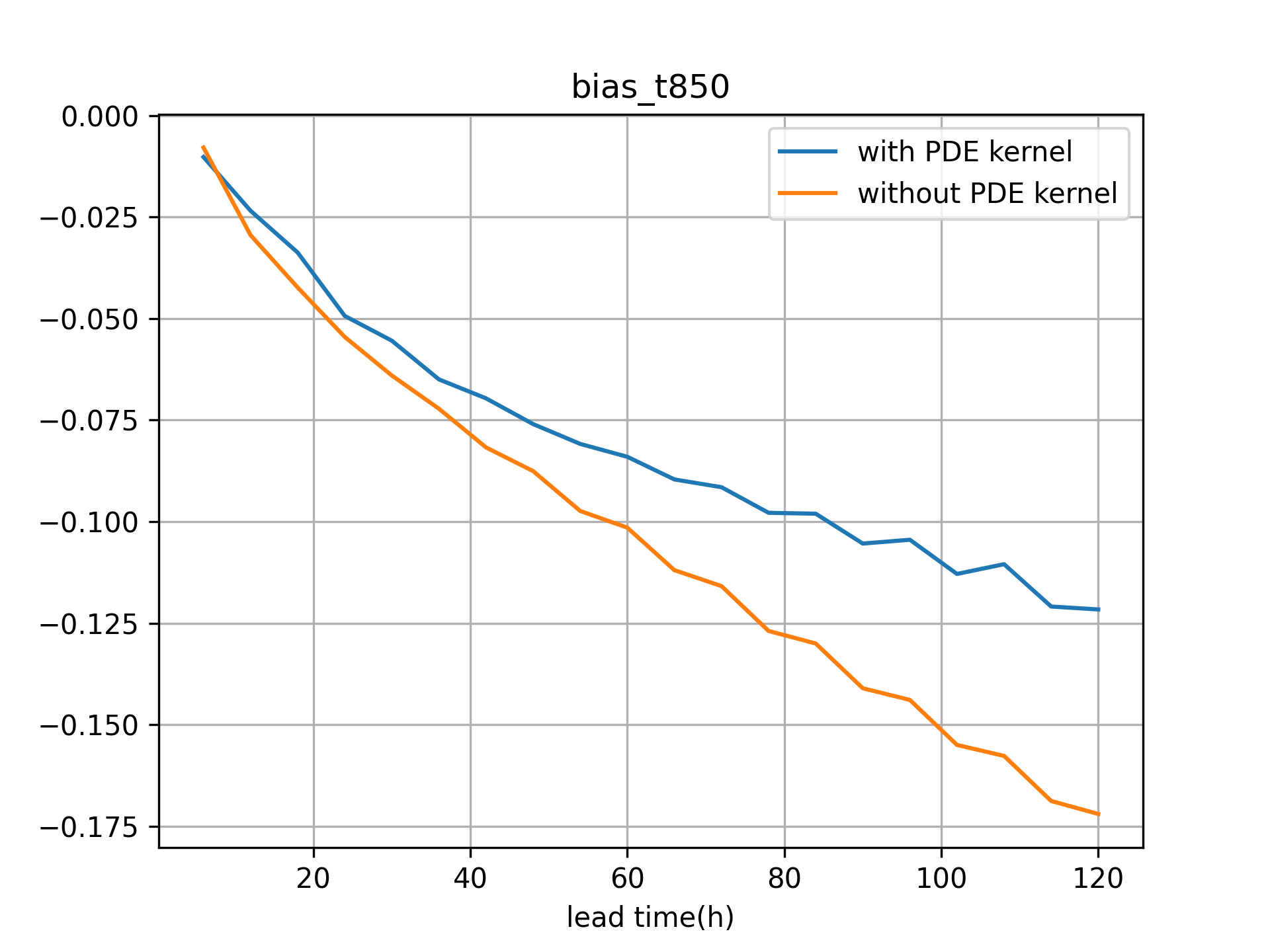}
    \includegraphics[width=4.5cm]{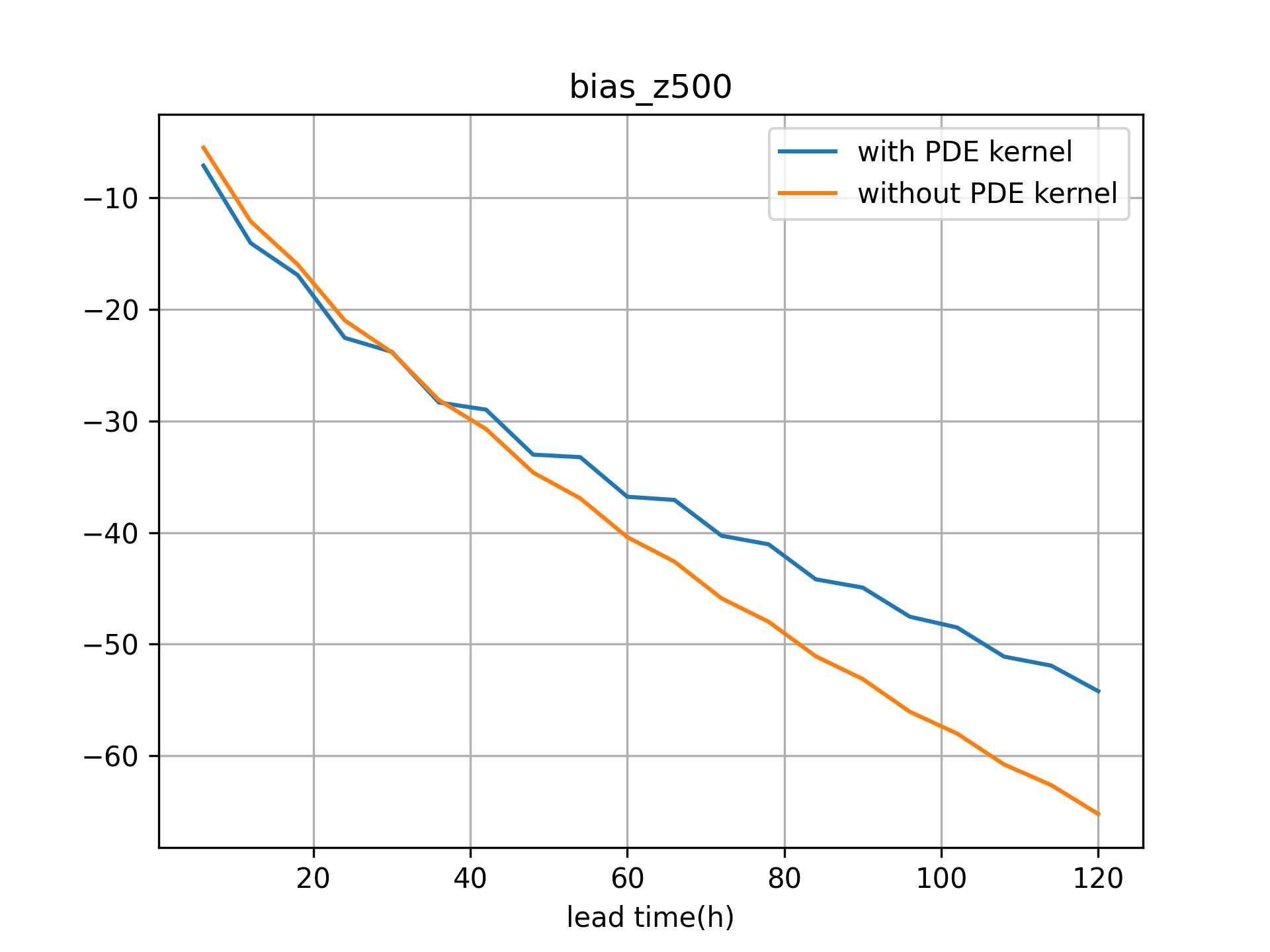}
    \caption{\textbf{Bias.} The closer to 0 the better.}
    \label{fig: bias}
\end{figure*}

\subsection{Prediction Energy Evaluation}

This assesses the energy~\cite{huo2023investigation} changes in the model's predictions. The experiments reveal that employing the PDE kernel aids in energy preservation.

\begin{equation}
\mathrm {energy} = \frac{1}{2} \left ( u^2+v^2 \right ) +\frac{c_p}{2T_r} T^2
\end{equation}


\begin{figure*}[!ht]
    \centering
    \includegraphics[width=4.5cm]{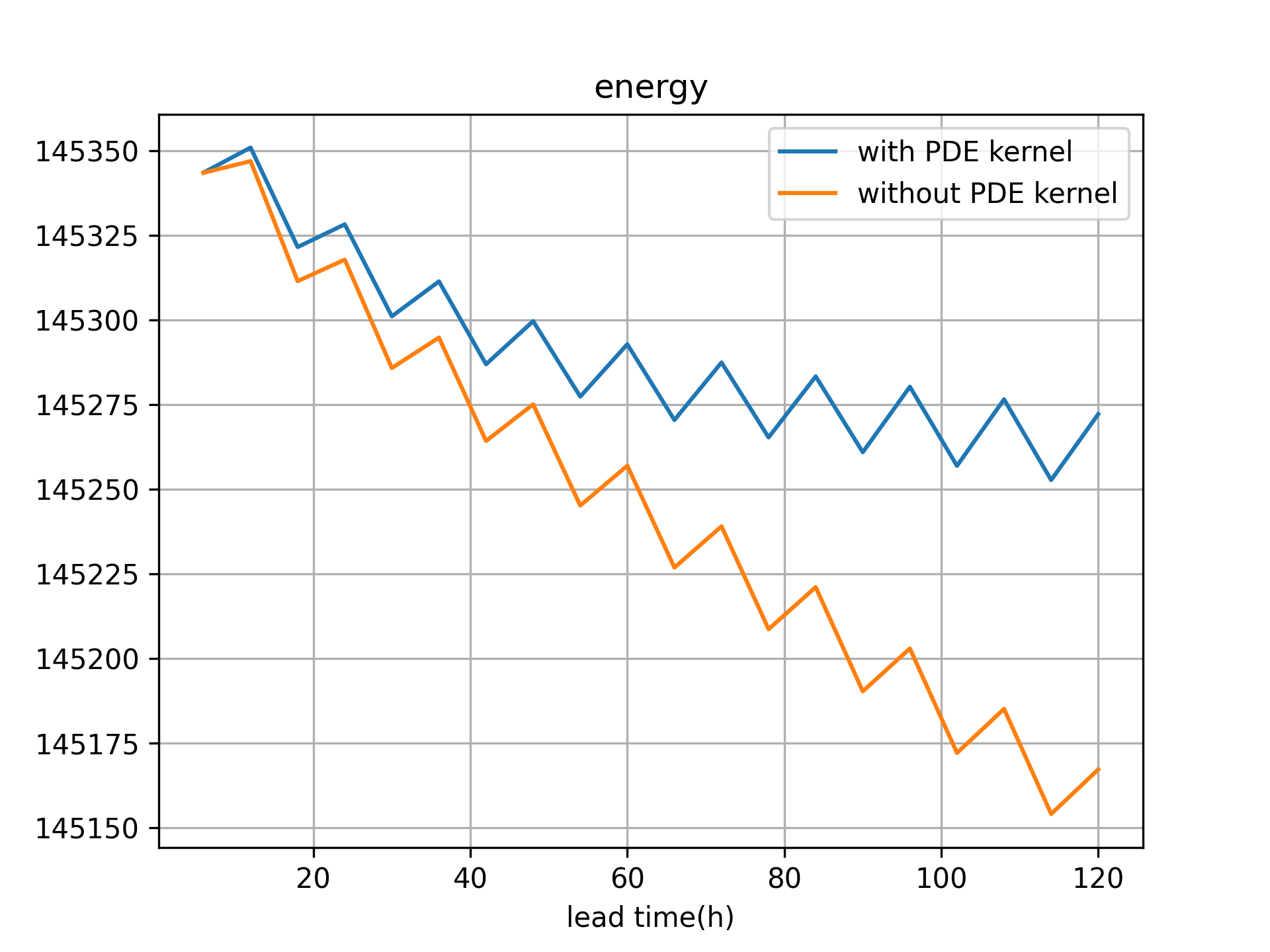} 
    \includegraphics[width=9cm]{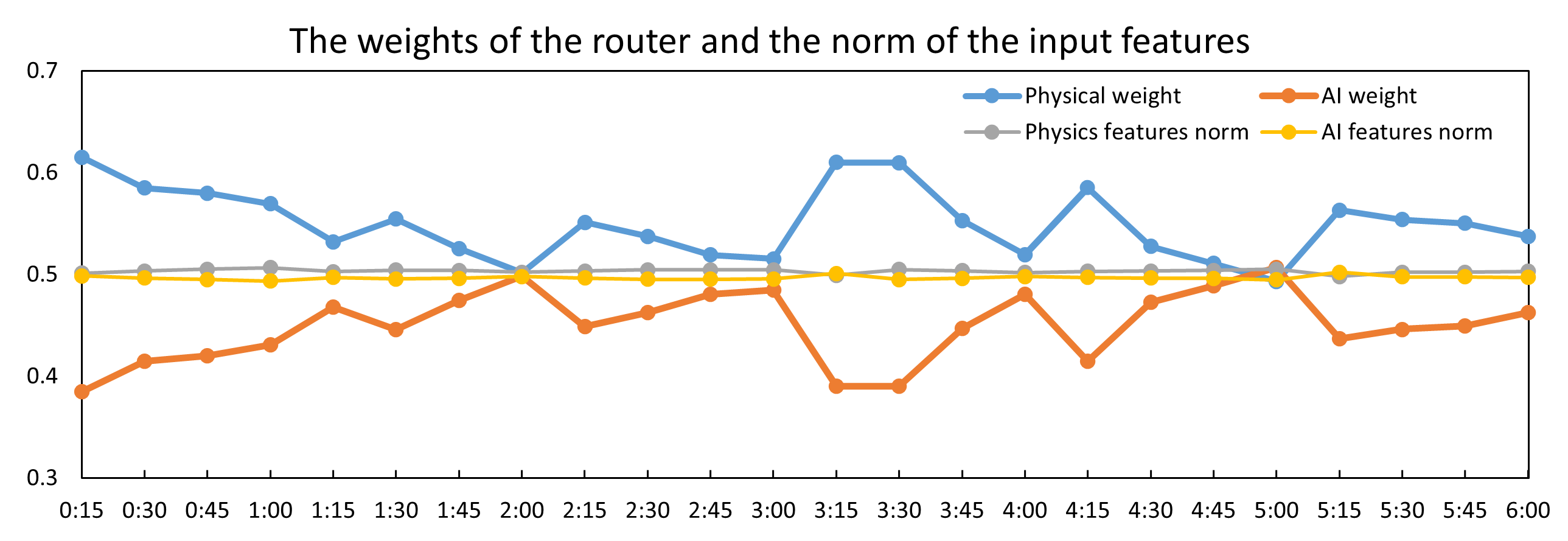}
    \caption{\textbf{Left.} Energy: the more consistent the better. \textbf{Right.} The norms of the outputs from the two networks are similar and stable. This indicates: a) The two networks produce outputs on the same scale. b) The router is decoupled and dynamically selects the more crucial features from the two branches without affecting the scale of the two networks.}
    \label{fig: Energy}
\end{figure*}

\subsection{Router Weights and Features Norm Change}

Figure \ref{fig: Energy} complements Figure \ref{fig: ratio} in the paper. It illustrates that physical and AI features are on a comparable scale, with the router dynamically selecting the more effective aspects from each. The router's weight adjustments do not impact the output of the AI or physical branches, highlighting the router's decoupling characteristics.

\section{Code Of Ethics and Broader Impacts}
\label{sec:Ethics}

Our research is ethical. The physical and AI hybrid model proposed in this paper can be used for global weather forecasting, which can serve many fields such as transportation and agriculture, and bring huge benefits to society.

The dataset used in this paper is public and there are no issues of infringement or privacy leakage. The experiments conducted in this paper are fair and reproducible. The resource consumption during the experiments is minimal and will not have an impact on the environment and society.

The model we propose is free of bias and discrimination issues. We open-source the model code and checkpoints on \href{https://github.com/black-yt/WeatherGFT}{GitHub}.

\section{Safeguard of Model}
\label{Safeguard}

This paper presents a hybrid physics-AI model for global weather forecasting. It is important to acknowledge that all models inherently carry a certain degree of forecasting error. Hence, the model proposed in this paper should not be solely relied upon as the sole basis for predicting significant events. Instead, it is recommended to integrate the findings from this model with other models and expert insights to draw comprehensive and informed conclusions.

\section{Assets}
\label{assets}

Our study adheres to the licenses governing the usage of existing assets, as the data utilized in this paper are publicly available and permitted for academic research purposes.

The model introduced in this paper represents a novel contribution and is considered a new asset.


\newpage
\section*{NeurIPS Paper Checklist}

\begin{enumerate}

\item {\bf Claims}
    \item[] Question: Do the main claims made in the abstract and introduction accurately reflect the paper's contributions and scope?
    \item[] Answer: \answerYes{} 
    \item[] Justification: We demonstrate the contribution and scope of the paper in the Abstract and Introduction \ref{sec:contributions}.

\item {\bf Limitations}
    \item[] Question: Does the paper discuss the limitations of the work performed by the authors?
    \item[] Answer: \answerYes{} 
        \item[] Justification: We show the limitations of our paper in Section \ref{sec: Conclusion}.

\item {\bf Theory Assumptions and Proofs}
    \item[] Question: For each theoretical result, does the paper provide the full set of assumptions and a complete (and correct) proof?
    \item[] Answer: \answerYes{} 
    \item[] Justification: We present the assumptions and proofs in Section \ref{sec:pdekernel} and Appendix \ref{pdes}.

    \item {\bf Experimental Result Reproducibility}
    \item[] Question: Does the paper fully disclose all the information needed to reproduce the main experimental results of the paper to the extent that it affects the main claims and/or conclusions of the paper (regardless of whether the code and data are provided or not)?
    \item[] Answer: \answerYes{} 
    \item[] Justification: We show the configuration required to reproduce the experiment in \ref{sec:ExperimentalSetup}. We show part of the model code in Appendix \ref{sec:code}, and the complete model code is released on \href{https://github.com/black-yt/WeatherGFT}{GitHub}.

\item {\bf Open access to data and code}
    \item[] Question: Does the paper provide open access to the data and code, with sufficient instructions to faithfully reproduce the main experimental results, as described in supplemental material?
    \item[] Answer: \answerYes{} 
    \item[] Justification: This paper uses the public datasets, there is no requirement to release the data. We show part of the model code in Appendix \ref{sec:code}, and the complete model code is released on \href{https://github.com/black-yt/WeatherGFT}{GitHub}.

\item {\bf Experimental Setting/Details}
    \item[] Question: Does the paper specify all the training and test details (e.g., data splits, hyperparameters, how they were chosen, type of optimizer, etc.) necessary to understand the results?
    \item[] Answer: \answerYes{} 
    \item[] Justification: We present the experimental setup and details in Section \ref{sec:ExperimentalSetup}.

\item {\bf Experiment Statistical Significance}
    \item[] Question: Does the paper report error bars suitably and correctly defined or other appropriate information about the statistical significance of the experiments?
    \item[] Answer: \answerYes{} 
    \item[] Justification: We show a detailed and correct error assessment in Section \ref{sec:mid} and Section \ref{sec:now}.

\item {\bf Experiments Compute Resources}
    \item[] Question: For each experiment, does the paper provide sufficient information on the computer resources (type of compute workers, memory, time of execution) needed to reproduce the experiments?
    \item[] Answer: \answerYes{} 
    \item[] Justification: We present the computational resources of our experiments in Section \ref{sec:ExperimentalSetup}.
    
\item {\bf Code Of Ethics}
    \item[] Question: Does the research conducted in the paper conform, in every respect, with the NeurIPS Code of Ethics \url{https://neurips.cc/public/EthicsGuidelines}?
    \item[] Answer: \answerYes{} 
    \item[] Justification: We explain in Appendix \ref{sec:Ethics} that our research is ethical.

\item {\bf Broader Impacts}
    \item[] Question: Does the paper discuss both potential positive societal impacts and negative societal impacts of the work performed?
    \item[] Answer: \answerYes{} 
    \item[] Justification: We show Broader Impacts in Appendix \ref{sec:Ethics}.
    
\item {\bf Safeguards}
    \item[] Question: Does the paper describe safeguards that have been put in place for responsible release of data or models that have a high risk for misuse (e.g., pretrained language models, image generators, or scraped datasets)?
    \item[] Answer: \answerYes{} 
    \item[] Justification: We explain the safeguards of the model in Appendix \ref{Safeguard}.

\item {\bf Licenses for existing assets}
    \item[] Question: Are the creators or original owners of assets (e.g., code, data, models), used in the paper, properly credited and are the license and terms of use explicitly mentioned and properly respected?
    \item[] Answer: \answerYes{} 
    \item[] Justification: We explain the usage of the existing assets in Appendix \ref{assets}.

\item {\bf New Assets}
    \item[] Question: Are new assets introduced in the paper well documented and is the documentation provided alongside the assets?
    \item[] Answer: \answerYes{} 
    \item[] Justification: We explain the new assets in Appendix \ref{assets}.

\item {\bf Crowdsourcing and Research with Human Subjects}
    \item[] Question: For crowdsourcing experiments and research with human subjects, does the paper include the full text of instructions given to participants and screenshots, if applicable, as well as details about compensation (if any)? 
    \item[] Answer: \answerNA{} 
    \item[] Justification: This study does not include crowdsourcing experiments and research with human subjects.

\item {\bf Institutional Review Board (IRB) Approvals or Equivalent for Research with Human Subjects}
    \item[] Question: Does the paper describe potential risks incurred by study participants, whether such risks were disclosed to the subjects, and whether Institutional Review Board (IRB) approvals (or an equivalent approval/review based on the requirements of your country or institution) were obtained?
    \item[] Answer: \answerNA{} 
    \item[] Justification: This research is not related to Institutional Review Board (IRB) Approvals or Equivalent for Research with Human Subjects.

\end{enumerate}

\end{document}